\documentclass[12pt]{article}

\usepackage{newtxtext,newtxmath}

\usepackage{graphicx}

\usepackage[letterpaper,margin=1in]{geometry}

\renewenvironment{abstract}
	{\quotation}
	{\endquotation}

\date{}

\renewcommand\refname{References and Notes}

\makeatletter
\renewcommand{\fnum@figure}{\textbf{Figure \thefigure}}
\renewcommand{\fnum@table}{\textbf{Table \thetable}}
\makeatother

\usepackage{scicite}

\usepackage{url}
\usepackage{subcaption}

\def\scititle{
	The geomagnetic storm and Kp prediction using Wasserstein transformer
}
\title{\bfseries \boldmath \scititle}

\author{
	Beibei Li$^{1\ast\dagger}$
	\small$^{1}$Deep Space Exploration Lab, Hefei \& 230000, China.\and
	\small$^\ast$Corresponding author. Email: blb0607@gmail.com\and
}

\begin{document} 

\maketitle

\begin{abstract} \bfseries \boldmath
The accurate forecasting of geomagnetic activity is important. In this work, we present a novel multimodal Transformer based framework for predicting the 3 days and 5 days planetary Kp index by integrating heterogeneous data sources, including satellite measurements, solar images, and KP time series. A key innovation is the incorporation of the Wasserstein distance into the transformer and the loss function to align the probability distributions across modalities. Comparative experiments with the NOAA model demonstrate performance, accurately capturing both the quiet and storm phases of geomagnetic activity. This study underscores the potential of integrating machine learning techniques with traditional models for improved real time forecasting.
\end{abstract}

\section{Introduction}\label{sec1}

Geomagnetic storms are significant disturbances in Earth's magnetosphere caused by solar wind conditions, such as coronal mass ejections and high-speed solar wind streams originating from coronal holes. These storms can have profound effects on technological systems, including power grids, communication networks, and satellite operations. The Kp index is a crucial parameter for quantifying the intensity of these geomagnetic disturbances. This introduce the physical and mathematical mechanisms that drive geomagnetic storms and influence the Kp index. This mainly uses neural network to predict 3 days Kp index and compare with NASA product.

Geomagnetic storms result from a sequence of physical processes initiated by solar activity that culminate in significant disturbances in Earth's magnetic environment. Broadly, this formation process can be categorized into three stages: solar activity sources, the interaction between the solar wind and Earth's magnetic field, and the subsequent energy transfer and diffusion within the magnetosphere.

\subsection{Geomagnetic Storm Formation Process}

Geomagnetic storms\cite{gonzalez1999interplanetary} are driven by a sequence of physical processes that begin with solar activity and culminate in significant disturbances in Earth's magnetic environment. The formation process can be categorized into three primary stages:

\subsubsection{Solar Activity Sources}
Solar activity is the primary driver of geomagnetic storms. CMEs are large expulsions of plasma\cite{baumjohann1996basic} and magnetic field from the Sun's corona; they occur when magnetic instabilities eject massive amounts of charged particles and magnetic structures into space at speeds ranging from hundreds to thousands of kilometers per second. If directed toward Earth, a CME can interact with the magnetosphere and initiate a geomagnetic storm. In addition, high-speed solar wind streams originating from coronal holes or active regions on the Sun play a significant role. Coronal holes are regions in the corona that are less dense and cooler than their surroundings, characterized by open magnetic field lines that allow solar wind particles to escape more easily. These streams, which can persist for several solar rotations and are more common during certain phases of the solar cycle, carry dense, fast-moving charged particles and magnetic fields. When these high-speed streams collide with the slower solar wind ahead of them, shock waves are produced that compress Earth's magnetosphere, contributing to geomagnetic disturbances.

\subsubsection{Solar Wind and Earth's Magnetic Field Interaction}
Once solar wind conditions are established, the interaction between the solar wind and Earth's magnetic field becomes critical. Earth's magnetosphere acts as a protective bubble; however, when a high-speed, high-density solar wind stream reaches Earth, it exerts additional dynamic pressure, compressing the dayside boundary inward. This compression increases the density and strength of the magnetic field lines, setting the stage for magnetic reconnection. In this process, oppositely directed magnetic field lines break and reconnect, releasing vast amounts of energy. Particularly when the interplanetary magnetic field has a southward component, it is antiparallel to Earth's magnetic field at the magnetopause or in the magnetotail, thereby facilitating reconnection. This mechanism allows solar wind energy and particles to enter the magnetosphere, significantly increasing the ring current and intensifying geomagnetic disturbances\cite{watari2017geomagnetic} .

\subsubsection{Energy Transfer and Diffusion within the Magnetosphere}
Following the initial interaction, energy is transferred and diffused within the magnetosphere through several processes \cite{akasofu1981energy}. High energy ions injected via magnetic reconnection circulate around Earth, forming a ring current. This ring current, which comprises ions with energies ranging from tens to hundreds of keV moving in closed trajectories, induces a magnetic field that opposes Earth's main magnetic field, leading to a measurable decrease in ground-level magnetic field strength. Additionally, auroral activity occurs when electrons and protons accelerated by the ring current precipitate into the ionosphere, exciting nitrogen and oxygen atoms that emit light. This auroral activity serves as both a visual indicator of geomagnetic storms and a contributor to ionospheric disturbances that can affect communication systems. Energy input from the ring current and auroral processes also increases ionization levels and alters the conductivity of the ionosphere, potentially degrading the performance of high frequency radio communications and global navigation satellite systems.

\subsection{Magnetic Reconnection}
Magnetic reconnection plays a pivotal role in the dynamics of geomagnetic storms by transferring energy from the solar wind into Earth's magnetosphere. Continuous solar wind pressure stretches Earth's magnetic field lines, forming a magnetotail where the field lines become elongated and twisted conditions that favor reconnection. When the interplanetary magnetic field exhibits a strong southward component, it aligns antiparallel to Earth's magnetic field at the magnetopause, effectively lowering the energy barrier for reconnection. As a result, stored magnetic energy is rapidly converted into kinetic and thermal energy, accelerating charged particles along reconnected field lines. These high-energy particles are injected into near-Earth magnetospheric orbits, amplifying the ring current which perturbs the global magnetic field, causing observable decreases in ground-level magnetic field strength and contributing to an increase in the Kp index a measure of geomagnetic activity. Moreover, energy released locally in the magnetotail is transmitted along magnetic field lines to other regions of the magnetosphere, amplifying disturbances on a global scale. The entire process is underpinned by magnetohydrodynamic equations and is quantified by reconnection rate models that relate the rate to the product of the magnitude of and the solar wind speed.
\subsection{Ring Current, Earth's Currents, Magnetic Fields, and Energy Changes During Geomagnetic Storms}
The ring current is a key component in geomagnetic storms. It forms when magnetic reconnection accelerates charged particles mainly protons and some heavy ions to high energies. These particles are then injected into near Earth orbits, where they follow helical paths along magnetic field lines and create an electric current. This current produces a magnetic field that typically opposes Earth's main magnetic field, leading to a reduction in the overall magnetic field observed at the Earth's surface. As the strength of the ring current increases, the magnetic field disturbance grows, which is reflected by a higher Kp index. Observational data and empirical models confirm the close relationship between the intensity of the ring current\cite{dessler1959effect}, the resulting magnetic perturbations, and the Kp index.\\
Understanding the interplay between Earth's currents, magnetic fields, and energy distribution is crucial for comprehending the full impact of geomagnetic storms. Earth's magnetic field is generated by the geodynamo in its liquid outer core through convective motions of conductive materials such as iron and nickel, forming a dominant primary dipolar field. This main field is further modified by induced magnetic fields from Earth's conductive crust in response to external variations, as well as by external contributions from interactions with the solar wind, including inputs from the magnetosheath and auroral zones. During geomagnetic storms, several electric currents emerge that further perturb the magnetic field: the ring current by high-energy ion injection via magnetic reconnection circulates in the equatorial magnetosphere and produces a magnetic field that opposes Earth's main field; auroral electrojets, driven by precipitating high-energy particles, contribute localized magnetic disturbances near the auroral zones; and field line currents connect the magnetosphere to the ionosphere, facilitating energy and momentum transfer. Moreover, magnetic reconnection converts stored magnetic energy into kinetic and thermal energy, and the released energy is transmitted along magnetic field lines to various regions, including the dayside and ionosphere, thereby amplifying disturbances across the entire magnetosphere.

\subsection{Impact of Geomagnetic Storms on Earth's Magnetic Field and Energy Systems}
Geomagnetic storms can significantly impact both technological and natural systems by altering magnetic fields and energy distributions. On the technological side, these storms induce currents in power transmission lines and transformers, which can lead to overheating, equipment damage, and widespread power outages; they also cause ionospheric disturbances that disrupt HF radio propagation, resulting in communication blackouts and degraded signal quality for aviation and maritime operations, and increase atmospheric density at satellite altitudes, which raises orbital drag and exposes satellite electronics and solar panels to damaging particle fluxes. In natural systems auroral activity while primarily a visual phenomenon signals significant geomagnetic disturbances that can also disrupt animal navigation by interfering with species that rely on Earth's magnetic field for orientation. Moreover, energy released during geomagnetic storms is redistributed throughout the magnetosphere and ionosphere, altering the balance of energy flows and causing measurable variations in the ground-level magnetic field, changes that are quantified by indices such as Kp and are crucial for monitoring and forecasting geomagnetic activity.\\
The Kp-index is the global geomagnetic activity index that is based on 3-hour measurements from ground-based magnetometers around the world. The Estimated 3-hour Planetary Kp-index is a preliminary Kp-index derived at the NOAA Space Weather Prediction Center using minute by minute data from a number of ground-based magnetometers that relay data in near-real time. These observatories are located in the United States, Canada, the United Kingdom, Germany, and Australia. The Kp-index ranges from 0 to 9 where a value of 0 means that there is very little geomagnetic activity and a value of 9 means extreme geomagnetic storming. For further details on the NOAA scales and the planetary K-index, please refer to \cite{noaaScalesExplanation}.
\subsection{Machine learning prediction for geomagnetic storms and Kp index}
The accurate forecasting of the Kp geomagnetic index is crucial for monitoring space weather and mitigating its impacts on critical infrastructure, such as power grids, communication networks, and satellite operations. By reliably predicting fluctuations in Earth's magnetic field, stakeholders can implement timely protective measures and reduce the risks associated with geomagnetic storms.\\
Numerous recent breakthroughs in machine learning, now in its golden age, make it imperative to explore how these advances can benefit forecasting. This review \cite{camporeale2019challenge} surveys previous work in these areas and highlights open challenges, providing a gentle introduction to machine learning tailored for space weather forecasting. This ensemble learning framework, CMETNet, \cite{alobaid2022predicting} predicts the arrival time of Earth-directed CMEs by integrating CME features, solar wind parameters, and SOHO/LASCO C2 coronagraph images. This machine learning-based model for the next 3-day geomagnetic
index (Kp) forecast\cite{wang2023machine} proposes a 3-day Kp index forecast model that uses previous Kp time series and SDO/AIA 193 Å images to predict key geomagnetic storm parameters such as start time, maximum intensity, and duration. The deep learning-based global forecasting model\cite{upendran2023global} predicts the time variability of Earth's horizontal magnetic field perturbation (dB/dt) a proxy for Geomagnetically Induced Currents using only solar wind measurements. By employing a Gated Recurrent Unit to summarize solar wind data and predicting spherical harmonic coefficients for high-cadence global forecasts, the model delivers rapid and accurate predictions that outperform or match benchmark models. 
\\
\\
In this paper, we introduce a novel neural network approach for predicting the 3 day and 5 day Kp index by integrating multimodal satellite and solar data. Our comparative analysis with NASA's forecasts demonstrates that our model achieves performance comparable to current state-of-the-art products.

\section{Methods}

This section details a multi-stage approach to preprocess time series data, generate training samples via a sliding window with class expansion (data oversampling), and build a multi-input Transformer-based model that integrates Wasserstein distribution alignment and an auxiliary high KP branch for prediction.

\subsection{Data Acquisition and Preprocessing}
This image is provided by NASA’s Solar Dynamics Observatory using its Atmospheric Imaging Assembly at the 193 Å wavelength from \cite{sdoData} as following Figure~\ref{fig:2023}, \ref{fig:2024}. The satellite data is provided by \cite{omniweb}, which integrates measurements from multiple satellites. This comprehensive dataset includes information on solar wind, the interplanetary magnetic field, and various parameters of the near-Earth space environment. These observations are crucial for understanding the interactions between the sun and the Earth and for monitoring and forecasting.
The KP data and NOAA model data is from \cite{noaaKp}.
The dataset covers the period from January 1, 2022, to July 1, 2024, and includes more than 140,000 images.
 \begin{figure}[htbp]   
    \vspace{0.5em}
    \begin{subfigure}[b]{0.475\textwidth}
        \centering
        \includegraphics[width=\textwidth]{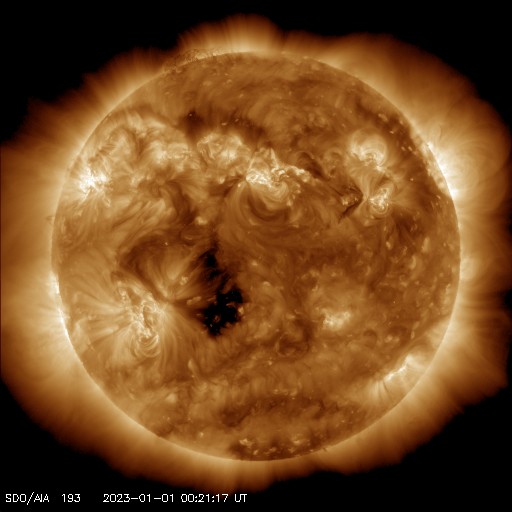}
        \caption{\textbf{The image on 20230101}}
        \label{fig:2023}
    \end{subfigure}
    \begin{subfigure}[b]{0.475\textwidth}
        \centering
        \includegraphics[width=\textwidth]{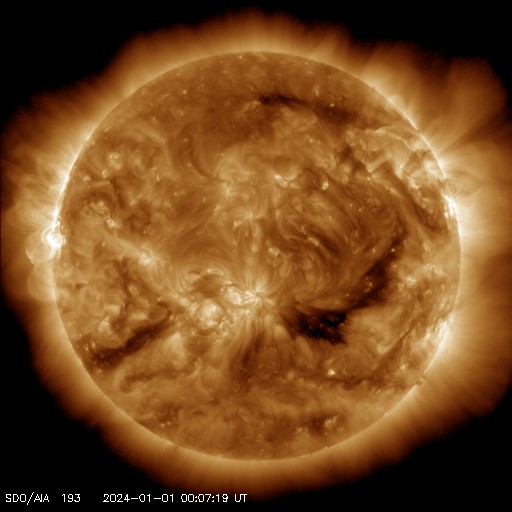}
        \caption{\textbf{The image on 20240101}}
        \label{fig:2024}
    \end{subfigure}
\end{figure}

\paragraph{Image Feature}

Image features are extracted using a pre-trained Swin Transformer\cite{liu2021swin} model. To accommodate larger images, the model configuration is updated accordingly. The process for each image is as follows:
\begin{itemize}
    \item Images are loaded at their original resolution and normalized. A quality check is performed by applying a ring-shaped mask that calculates the ratio of low-intensity pixels within a predefined annular region. The ring-shaped mask is applied, and the ratio of dark pixels with intensity below a threshold is computed. If the ratio exceeds 14\%, the image is discarded. 
    \item For images passing the quality check, the Swin Transformer extracts features. The last hidden state is averaged to produce a fixed-length feature vector. These features are stored in a dictionary keyed by the corresponding hour. 
\end{itemize}

\paragraph{Satellite Data Processing}
 In parallel, satellite data is preprocessed by merging date-related columns into a unified datetime index and resampling the data at an hourly frequency with forward filling to address missing values. 
 \begin{itemize}
    \item A subset of relevant features is retained for analysis and modeling. These include:
    \begin{itemize}
        \item \textbf{Temporal Features:} \texttt{YEAR}, \texttt{DOY}, \texttt{Hour}.
        \item \textbf{Magnetic Field Measurements:} \texttt{Scalar B, nT}, \texttt{Vector B Magnitude, nT}, \texttt{Lat. Angle of B (GSE)}, \texttt{Long. Angle of B (GSE)}, \texttt{BX, nT (GSE, GSM)}, \texttt{BY, nT (GSE)}, \texttt{BZ, nT (GSE)}, \texttt{BY, nT (GSM)}, \texttt{BZ, nT (GSM)}.
        \item \textbf{Derived Metrics:} \texttt{RMS\_magnitude, nT}, \texttt{RMS\_field\_vector, nT}, \texttt{RMS\_BX\_GSE, nT}, \texttt{RMS\_BY\_GSE, nT}, \texttt{RMS\_BZ\_GSE, nT}.
        \item \textbf{Plasma Parameters:} \texttt{SW Plasma Temperature, K}, \texttt{SW Proton Density, N/cm\textsuperscript{3}}, \texttt{SW Plasma Speed, km/s}, \texttt{SW Plasma flow long. angle}, \texttt{SW Plasma flow lat. angle}, \texttt{Alpha/Prot. ratio}, \texttt{Flow pressure}, \texttt{E elecrtic field}, \texttt{Plasma Beta}, \texttt{Alfen mach number}, \texttt{Magnetosonic Much num.}, \texttt{Quasy-Invariant}.
    \end{itemize}
\end{itemize}
Anomalous values are identified and replaced with \texttt{NaN} to remove spurious measurements. 
Missing values, now marked as \texttt{NaN}, are filled using linear interpolation.

\paragraph{KP Data}
The KP index, originally provided 3 hours frequency.
\paragraph{Integration of Datasets}
Finally, the satellite data, KP index data and image-derived features are merged based on their timestamps. 

\paragraph{Train and Test set Data Splitting}
The dataset covers the period from January 1, 2022, to July 1, 2024. The training set includes all data before April 20, 2024, while the test set spans from April 20, 2024, to July 1, 2024.

\subsection{Temporal Resampling}
To reduce noise and standardize the data frequency, all features are resampled into three-hour intervals. For each interval, the mean is computed for the image features, the numerical satellite measurements, and the KP index. Rows containing missing values are removed after resampling, yielding a complete and uniform dataset for further processing.

\subsection{Sliding Window Data Preparation and Data Expansion}
A sliding window approach is adopted to generate input-output pairs from the time series data:
A sliding window approach is employed to generate input-output pairs from the time series data. Two configurations are used: one for model training and another for daily forecasting.

\begin{itemize}
    \item \textbf{Windowing:} For training, an input window of 40 time steps (covering 5 days) is used to capture preceding temporal information, with an output window of 24 time steps (covering 3 days) or 40 time steps for prediction. For daily forecasting, an input window of 40 time steps is used while the output window is set to 24 time steps (covering 3 days) or 40 time steps.
    \item \textbf{Feature Transformation:} Image features are first standardized on a per-row basis and then reduced in dimensionality using Principal Component Analysis (PCA) to obtain a compact 512-dimensional representation. Numerical satellite features (excluding image features and temporal markers) are normalized column-wise.
    \item \textbf{Label Generation:} Three distinct KP index classification schemes are generated:
    \begin{enumerate}
        \item A fine-grained 28-class discretization.
        \item A 10-class categorization based on integer bins.
        \item A coarse 3-class categorization representing quiet, active, and storm conditions.
    \end{enumerate}
    Additionally, for each output time step a binary auxiliary label is created to indicate whether the KP index exceeds a predetermined high threshold (e.g., KP $\geq 7$).

    \item \textbf{Data Expansion:} To address class imbalance in the KP values, the method applies a data expansion strategy. For each KP value observed in the input windows, the number of samples is counted. The class with the highest frequency determines the target sample count, and for classes with fewer samples, additional instances are generated by randomly sampling from the existing instances until balanced with the most frequent class. These oversampled indices are then combined and shuffled to create a balanced training dataset.

\end{itemize}

\subsection{Model Architecture}
\begin{figure}[htbp]
    \centering
    \includegraphics[width=\textwidth]{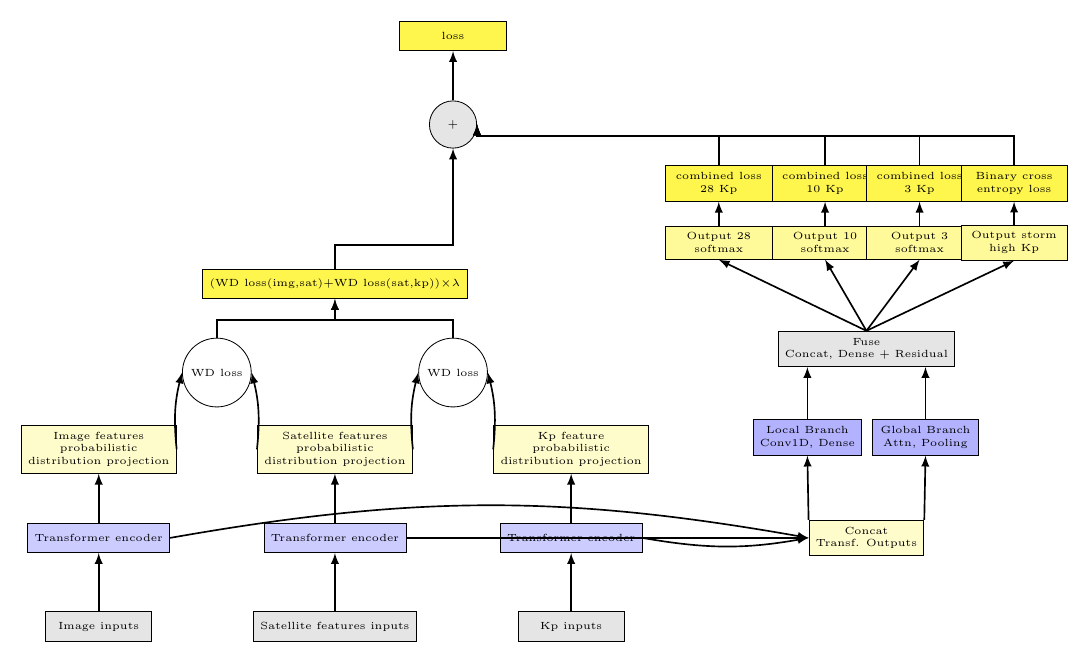}
    \caption{\textbf{The multimodal transformer architecture}}
    \label{fig:trans_arch}
\end{figure}

The model utilizes a multimodal architecture, see Figure~\ref{fig:trans_arch}, designed to integrate heterogeneous data sources:
\begin{enumerate}
    \item \textbf{Input Modalities:} Three separate inputs are considered:
    \begin{itemize}
        \item \textit{Image Features:} Pre-trained Swin Transformer to extract fixed-length feature vectors, row normed and PCA-reduced representations of image features.
        \item \textit{Satellite Features:} Normalized numerical satellite measurements.
        \item \textit{KP Index:} The normalized raw KP time series.
    \end{itemize}
    \item \textbf{Transformer Encoders:} Each modality is independently processed through a dedicated Transformer encoder block. Each encoder comprises a multi-head attention\cite{vaswani2017attention} layer, dropout for regularization, residual connections, and a feed-forward network with L2 regularization.
    \item \textbf{Projection and Distribution Alignment:} After the Transformer encoding, each branch is projected into a probability distribution over a fixed number of bins using a dense layer with softmax activation. To align the modalities, the Wasserstein distance is computed between the projected distributions. The Wasserstein distance, also known as the Earth Mover's Distance, measures the minimum cost required to transform one probability distribution into another. For two cumulative distribution functions (CDFs) \( F \) and \( G \), it is defined as:
    \[
    W(F, G) = \int_{-\infty}^{+\infty} \left| F(x) - G(x) \right| \, dx.
    \]    
    In our implementation, this is approximated by computing the absolute differences between the cumulative sums of the predicted and true distributions over the bins, and then taking the mean value of these differences. The resulting alignment loss is scaled by a hyperparameter and added to the total loss.
    \paragraph{Wasserstein Distance and Custom Loss}
The 1D discrete Wasserstein distance measures the minimal "effort" required to transform one probability distribution into another. Let 
    \[
    p = (p_1, p_2, \ldots, p_K) \quad \text{and} \quad q = (q_1, q_2, \ldots, q_K)
    \]
    be two probability distributions over $\{1,2,\ldots,K\}$. Their cumulative distribution functions are defined as:
    \[
    F_p(k) = \sum_{i=1}^{k} p_i, \quad F_q(k) = \sum_{i=1}^{k} q_i.
    \]
    The Wasserstein distance is then given by:
    \[
    W_1(p,q) = \sum_{k=1}^{K} \bigl\lvert F_p(k) - F_q(k) \bigr\rvert.
    \]
    In our model, this distance is computed between the projected distributions of different modalities. The resulting Wasserstein alignment loss is scaled by a hyperparameter and added to the total loss to encourage consistency across modalities.

    \item \textbf{Fusion and Output Layers:} The outputs from all three branches are concatenated. A local branch (using 1D convolution with residual connections) captures short-term temporal features, while a global branch (using attention and global average pooling) integrates long-term information. The fused features are then mapped to four distinct output branches:
    \begin{itemize}
        \item A 28-class classification branch (softmax).
        \item A 10-class classification branch (softmax).
        \item A 3-class classification branch (softmax).
        \item A binary auxiliary branch for high KP detection (sigmoid).
    \end{itemize}
\end{enumerate}

\subsection{Model Training}
The model is compiled with the Adam optimizer. For the multi-class outputs, a composite loss function is used that combines categorical cross-entropy and the scaled Wasserstein alignment loss which is called combined loss in Figure~\ref{fig:trans_arch}:
\[
L_{\text{combined}} = \alpha \cdot L_{\text{CE}} + (1-\alpha) \cdot L_{\text{Wass}},
\]
where $\alpha \in [0,1]$ is a hyperparameter balancing the two loss components, the $L_{\text{CE}} $ is the cross entropy categorical loss and $L_{\text{Wass}}$ is the Wasserstein Distance and Custom Loss. The binary high KP output is optimized using binary cross-entropy loss. Training strategies include early stopping, learning rate scheduling, and mini-batch training with a validation split. 
\\
There are two part of the total loss, one is that after the Transformer encoding, each branch is projected into a probability distribution over a fixed number of bins using a dense layer with softmax activation. To align the modalities, the Wasserstein distance is computed between the projected distributions as the loss. And the other one is the scaled Wasserstein alignment loss which is the cross entropy loss for the 28-class, 10-class, 3-class, and binary high KP plus the Wasserstein Distance and Custom Loss for the 28-class, 10-class, 3-class.

\subsection{Fine Tuning and Multi day Prediction Strategy}

The forecasting model is updated continuously using an online fine-tuning process that adapts to the most recent historical data while ensuring that no future information contaminates the training process. This strategy is composed of two integrated components: online fine-tuning and multi-day prediction.

\subsubsection*{Fine Tuning}
For each prediction day, the model is refined using a sliding window of past observations. This step is crucial for capturing local temporal dynamics and involves:
\begin{enumerate}
    \item The model is fine-tuned on a daily sample derived from a fixed-length sliding window that contains only past data. This practice guarantees that only historical information is used, completely avoiding data leakage.
    \item Each day's fine-tuning is performed for a predetermined number of epochs using the most recent available labeled data, ensuring the model stays updated with the latest trends.
\end{enumerate}

\subsubsection*{Multi day Prediction}
After fine-tuning, the model is employed to forecast future Kp values across multiple horizons. The multi-day prediction component is executed as follows:
\begin{enumerate}
    \item A sliding window is applied to the test dataset to partition it into daily samples. Each sample contains historical observations up to the current prediction day.
    \item The model has been fine-tuned on the current day's sample, subsequent daily samples are reserved exclusively for prediction:
    \begin{itemize}
        \item The sample immediately following the fine-tuning day is used for a 1-day forecast (24 hours later).
        \item The sample two days later is used for a 2-day forecast (48 hours later).
        \item Similarly, samples three, four, and five days ahead are cropped for 3-day, 4-day, and 5-day predictions, respectively.
    \end{itemize}
\end{enumerate}

\textbf{Prevention of Data Leakage:}  
The entire procedure is meticulously designed to prevent any contamination from future data:
\begin{itemize}
    \item All data transformations (e.g., normalization, PCA, scaling) are performed using parameters derived solely from the training data.
    \item The fine-tuning and the prediction steps utilize only historical information.
\end{itemize}

In summary, by combining online fine-tuning with a systematic multi-day prediction strategy, the method achieves robust and unbiased forecasting of the Kp geomagnetic index. The model is continuously updated with the most recent labeled data while future predictions for horizons ranging from 1 to 3 days or 1 to 5 days are generated using carefully cropped daily samples.


\section{Results}
\begin{figure}[htbp]
    \centering
    \includegraphics[width=\textwidth]{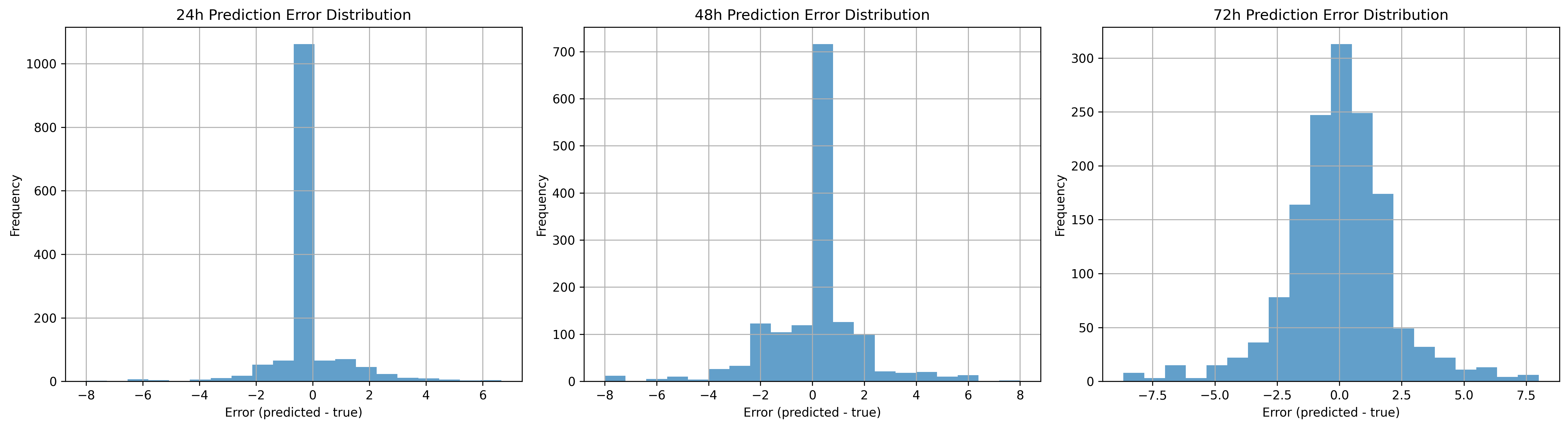}
    \caption{\textbf{The multimodal model error distribution}}
    \label{fig:prediction_error_distribution}
\end{figure}

\begin{figure}[htbp]
    \centering
    \includegraphics[width=\textwidth]{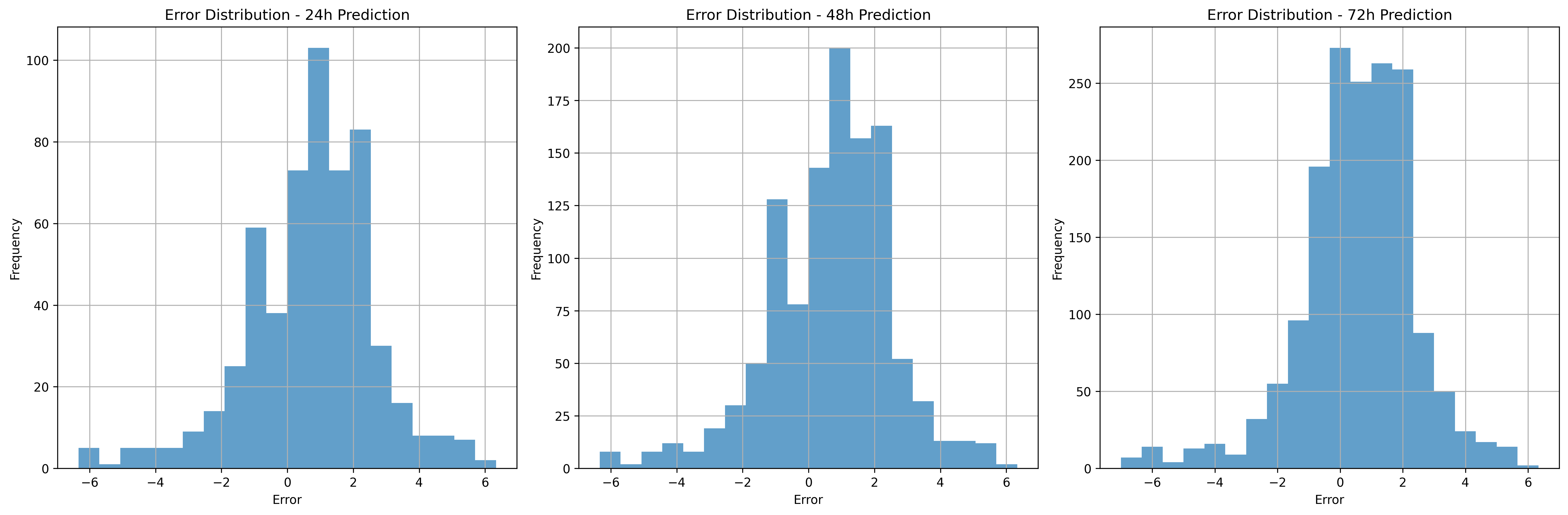}
    \caption{\textbf{The NOAA model error distribution}}
    \label{fig:error_distribution_combined}
\end{figure}

Here are the error distributions for one day prediction, two day prediction, and three day prediction from the multimodal model \ref{fig:prediction_error_distribution} and NOAA model \ref{fig:error_distribution_combined}.


\begin{figure}[htbp]
    \centering
    \begin{subfigure}[b]{0.475\textwidth}
        \centering
        \includegraphics[width=\textwidth]{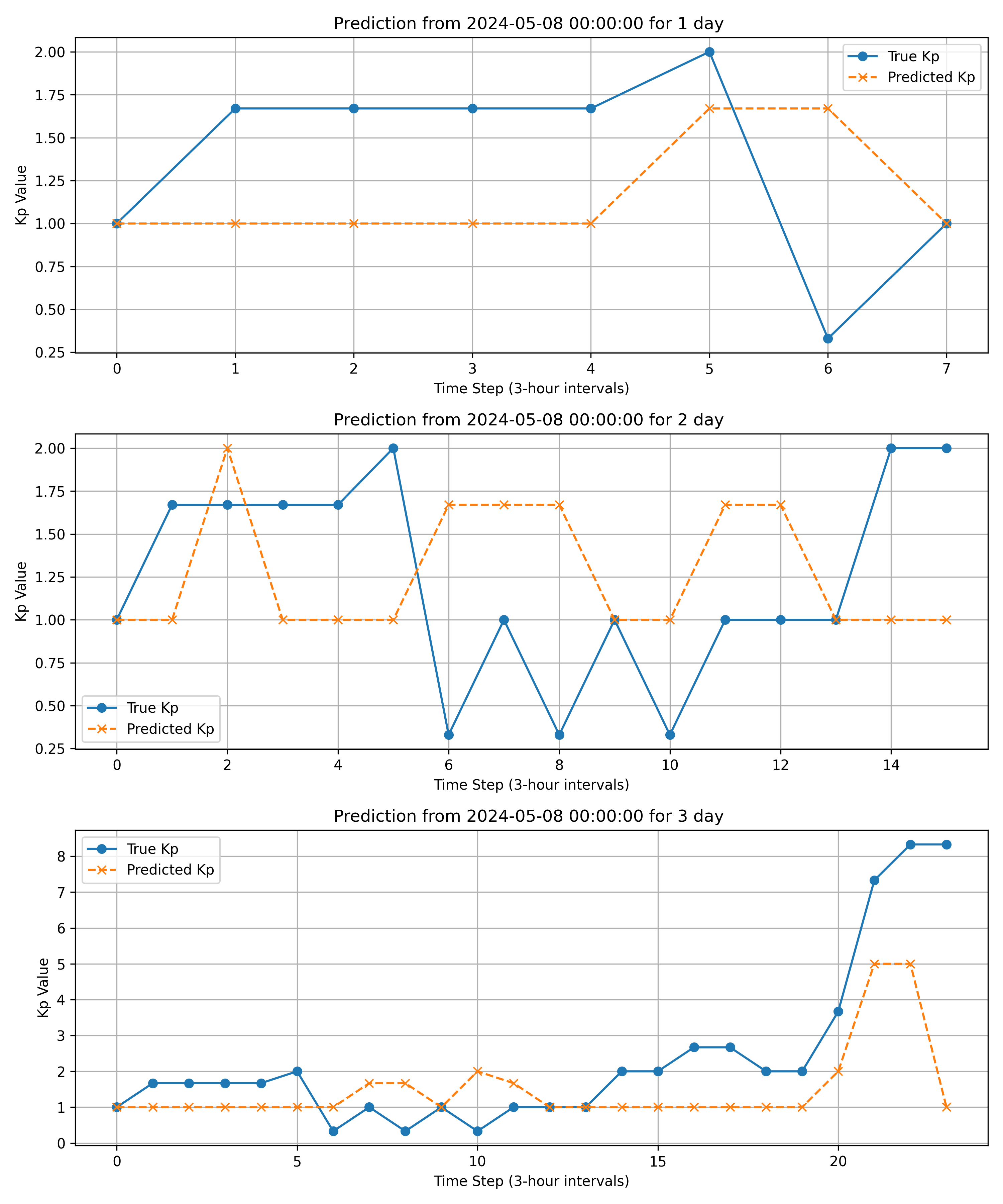}
        \caption{May 8}
        \label{fig:prediction_2024-05-08}
    \end{subfigure}
    \begin{subfigure}[b]{0.475\textwidth}
        \centering
        \includegraphics[width=\textwidth]{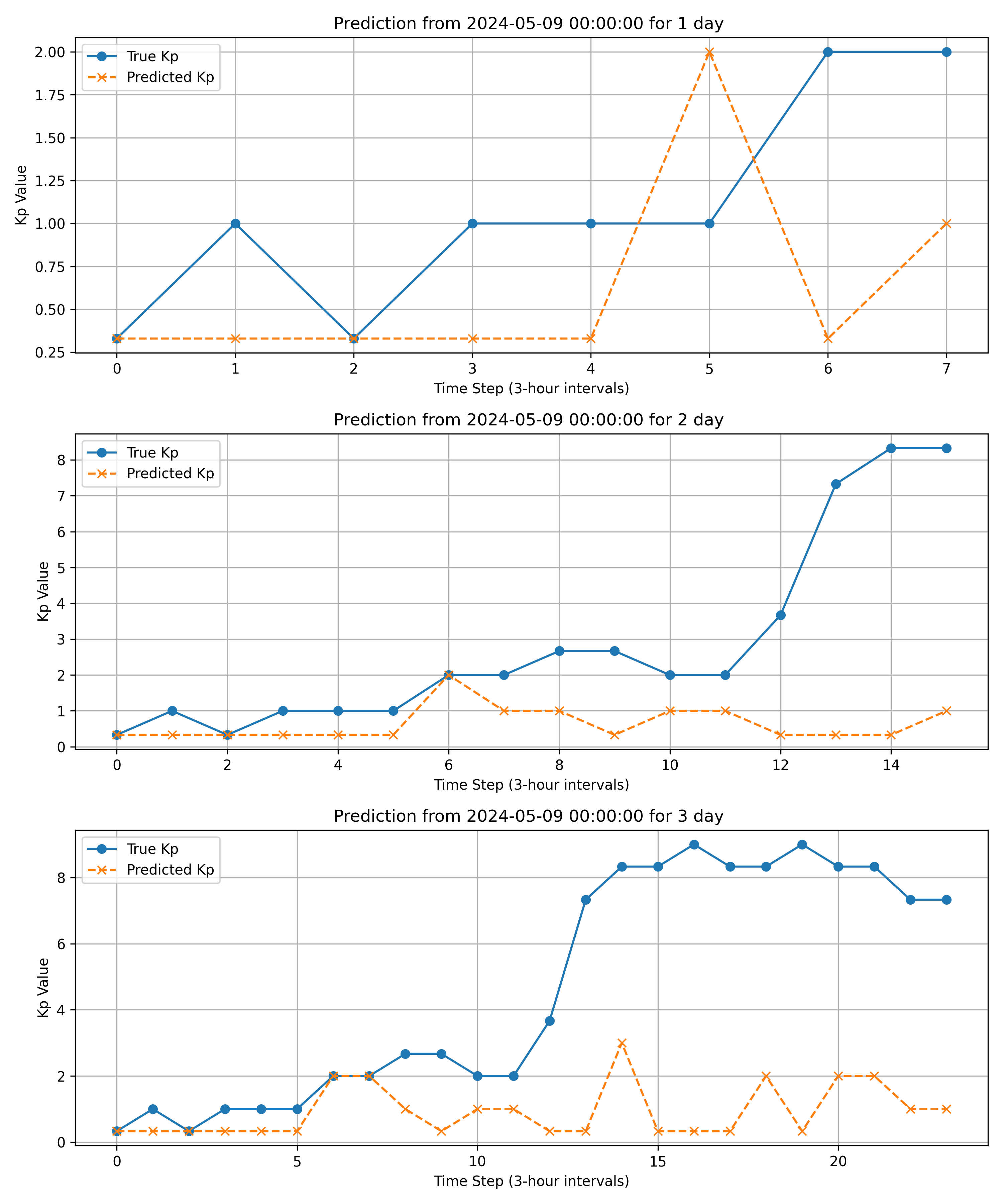}
        \caption{May 9}
        \label{fig:prediction_2024-05-09}
    \end{subfigure}

    \vspace{0.5em}
    \begin{subfigure}[b]{0.475\textwidth}
        \centering
        \includegraphics[width=\textwidth]{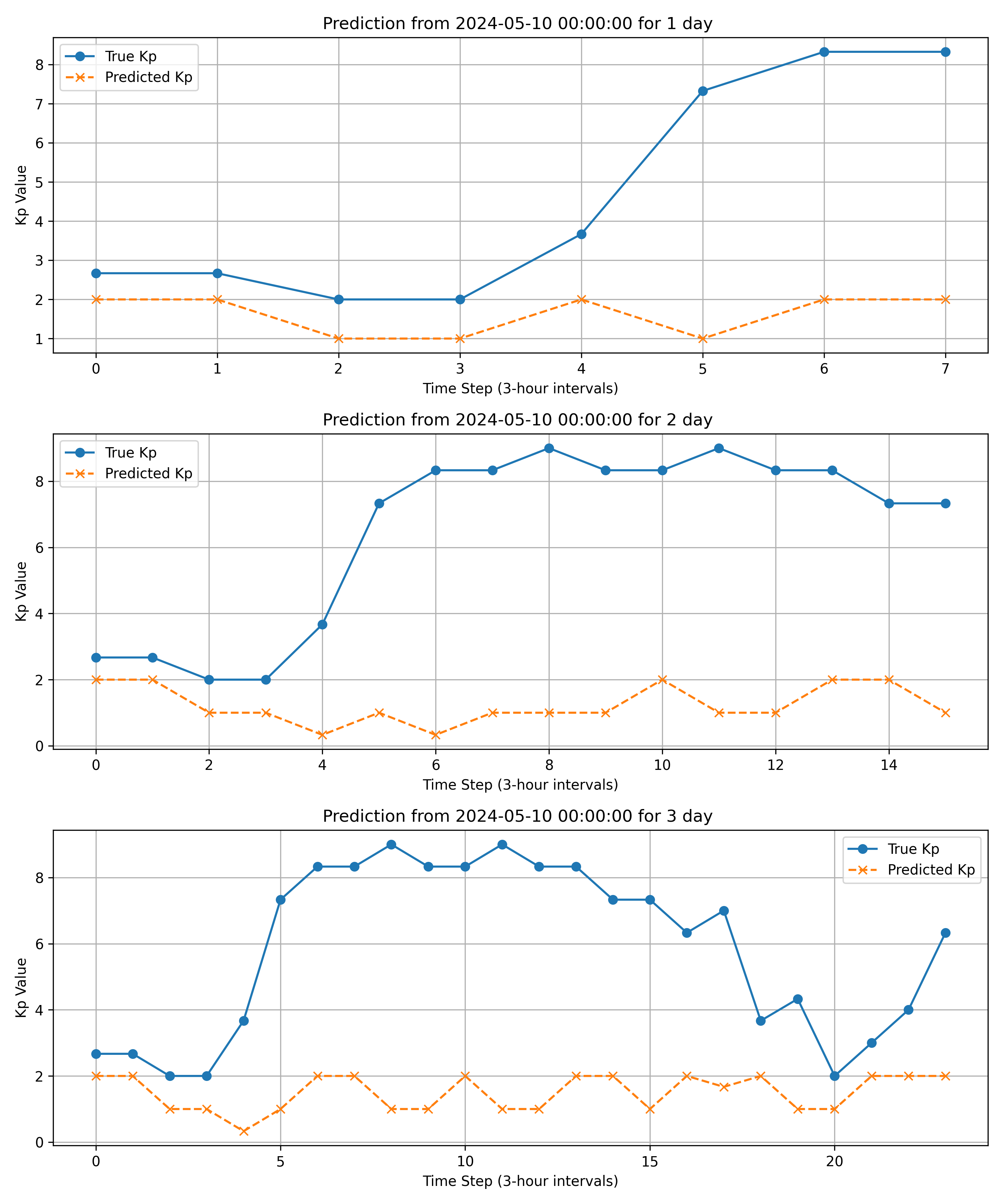}
        \caption{May 10}
        \label{fig:prediction_2024-05-10}
    \end{subfigure}
    \begin{subfigure}[b]{0.475\textwidth}
        \centering
        \includegraphics[width=\textwidth]{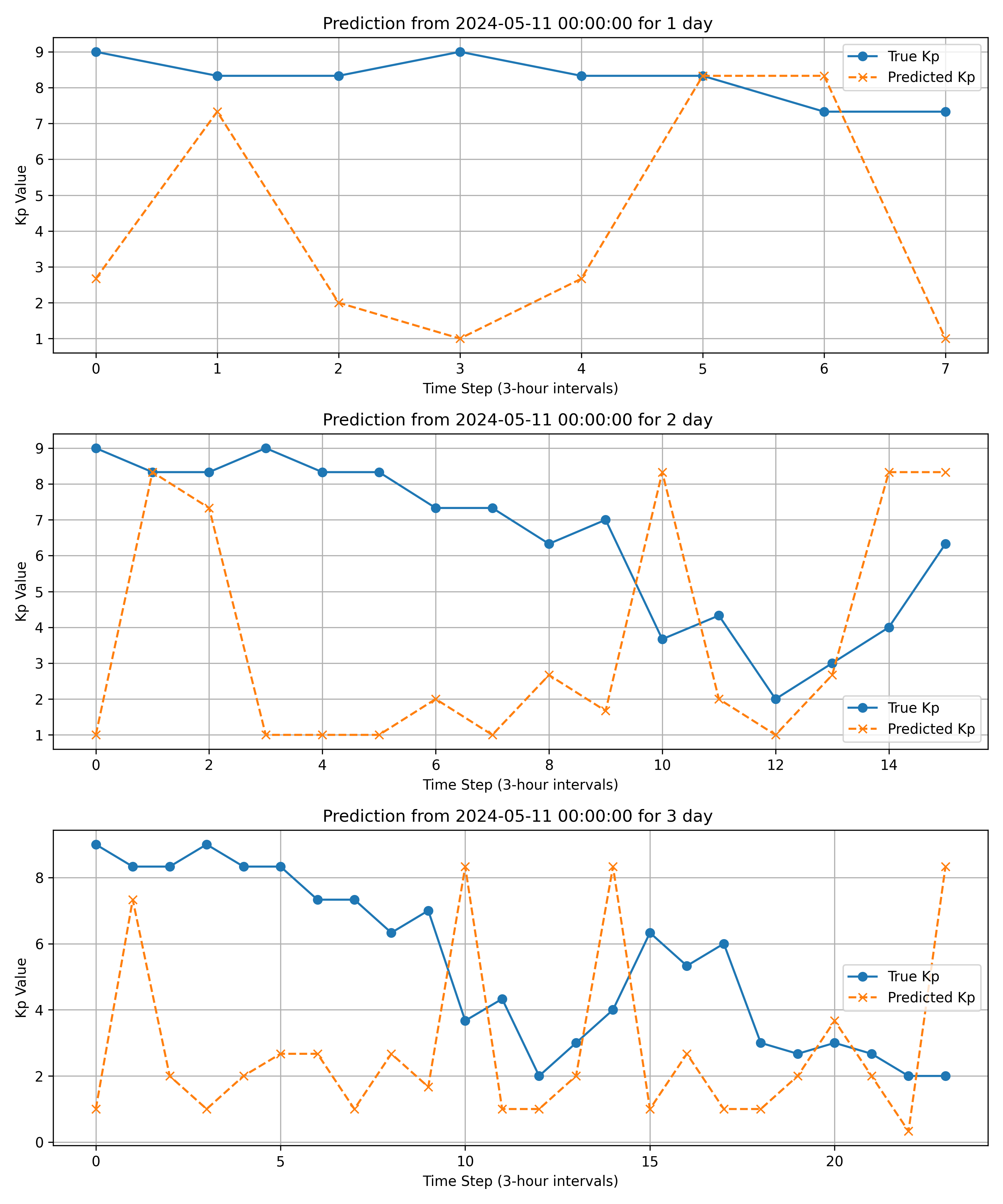}
        \caption{May 11}
        \label{fig:prediction_2024-05-11}
    \end{subfigure}
    
    \caption{\textbf{The multimodal model prediction results from May 8 to May 11}}
    \label{fig:predictions_8}
 \end{figure}
\begin{figure}[htbp]   
    \vspace{0.5em}
    \begin{subfigure}[b]{0.475\textwidth}
        \centering
        \includegraphics[width=\textwidth]{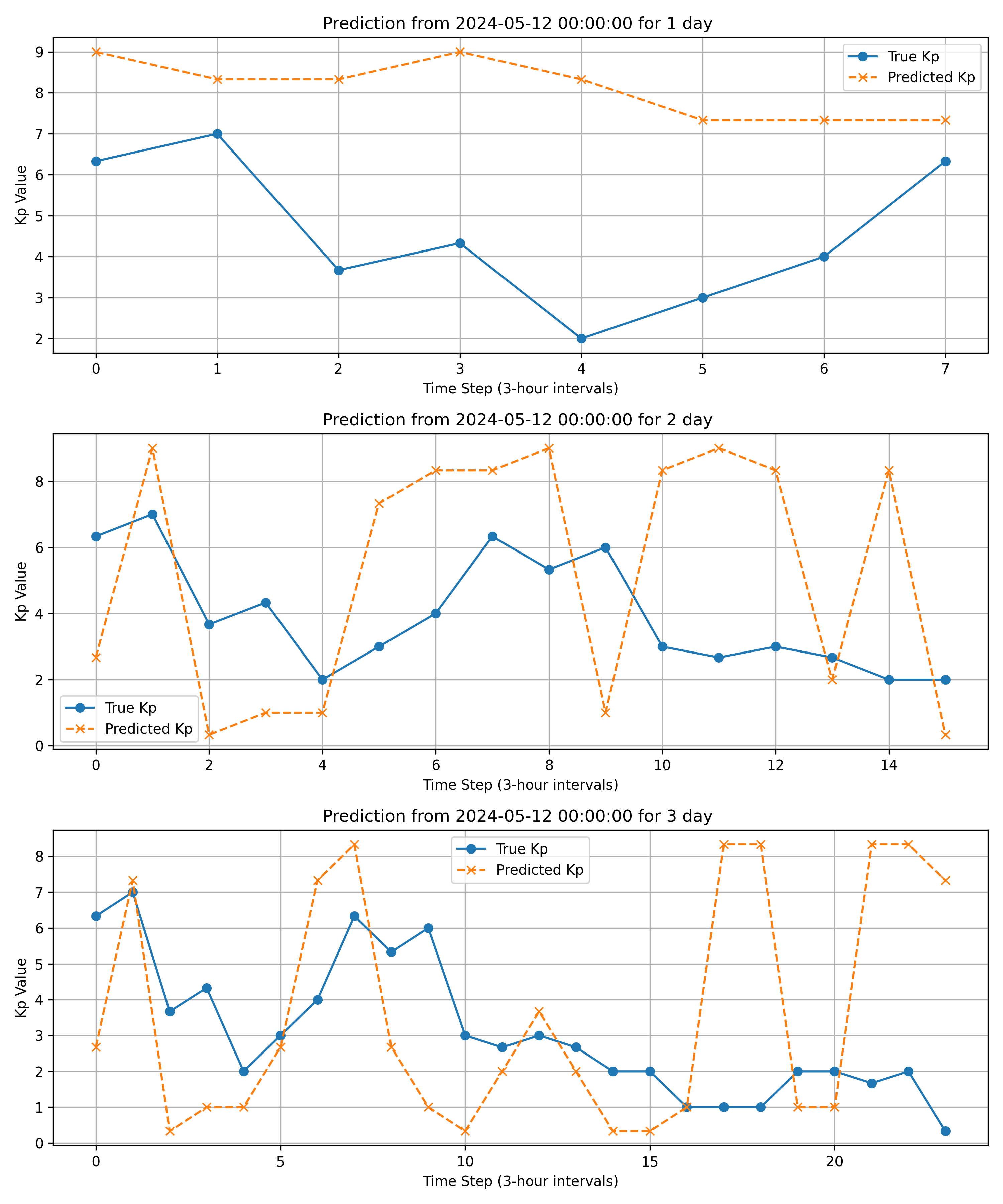}
        \caption{May 12}
        \label{fig:prediction_2024-05-12}
    \end{subfigure}
    \begin{subfigure}[b]{0.475\textwidth}
        \centering
        \includegraphics[width=\textwidth]{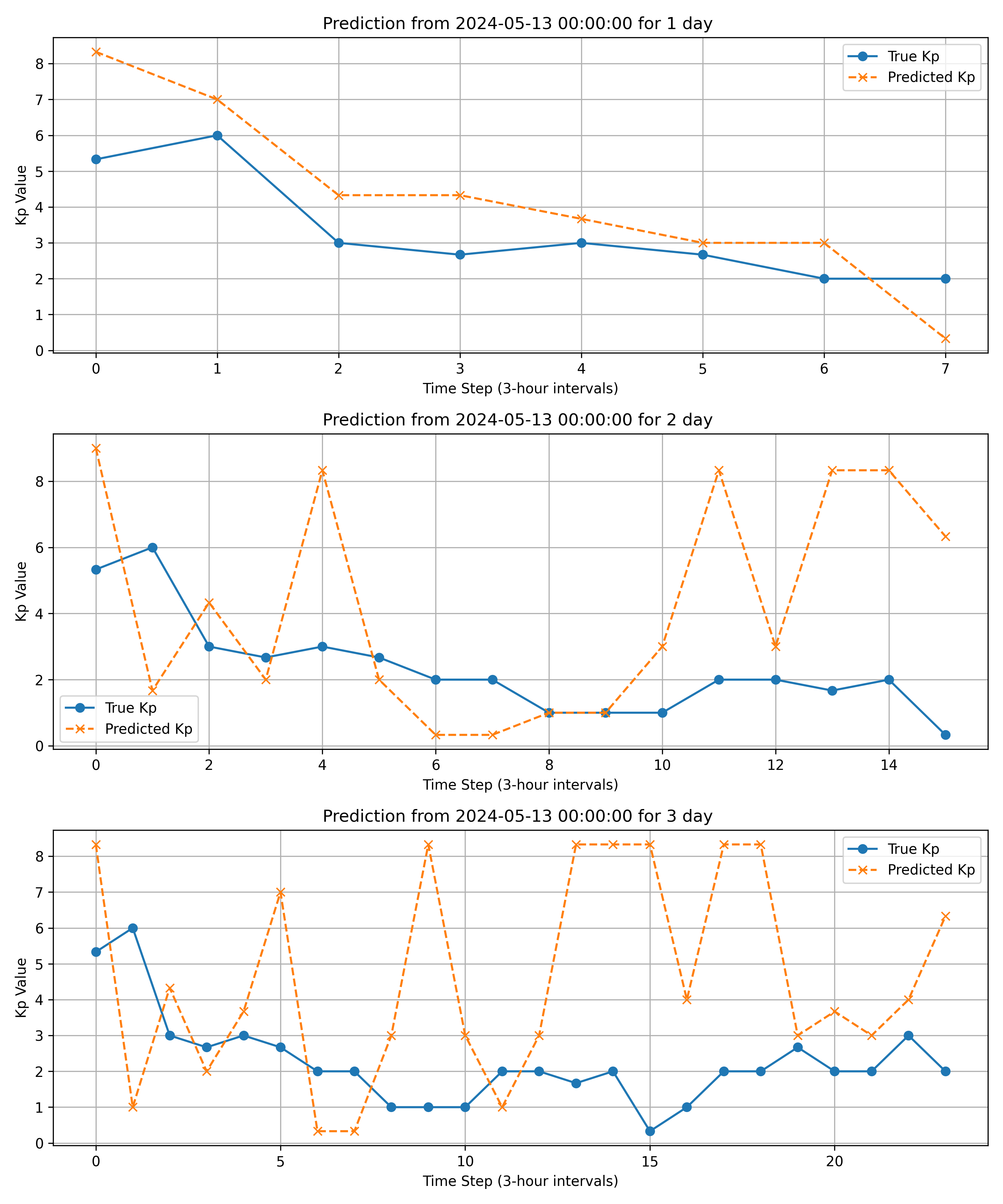}
        \caption{May 13}
        \label{fig:prediction_2024-05-13}
    \end{subfigure}

    \vspace{0.5em}
    \begin{subfigure}[b]{0.475\textwidth}
        \centering
        \includegraphics[width=\textwidth]{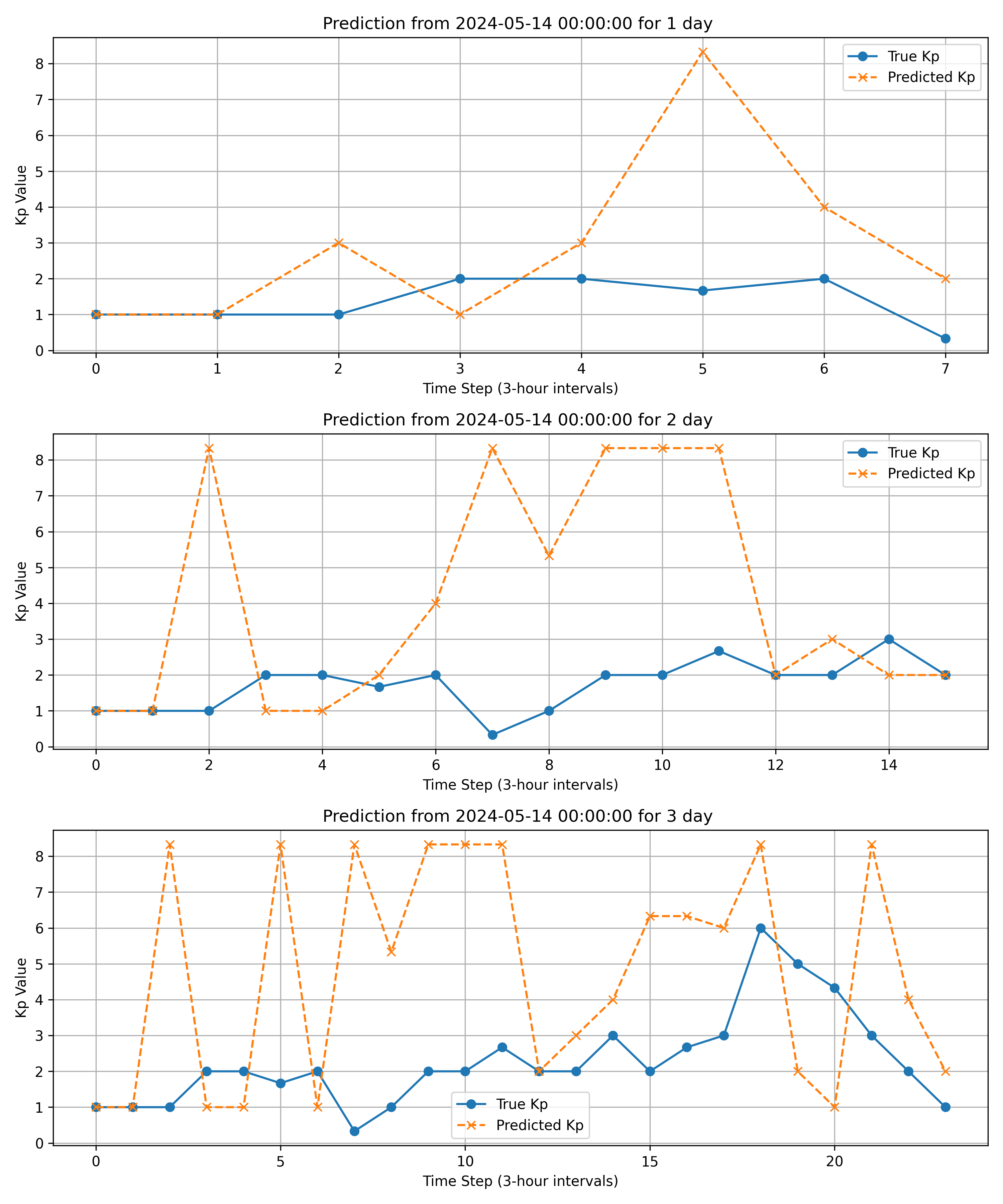}
        \caption{May 14}
        \label{fig:prediction_2024-05-14}
    \end{subfigure}
    \begin{subfigure}[b]{0.475\textwidth}
        \centering
        \includegraphics[width=\textwidth]{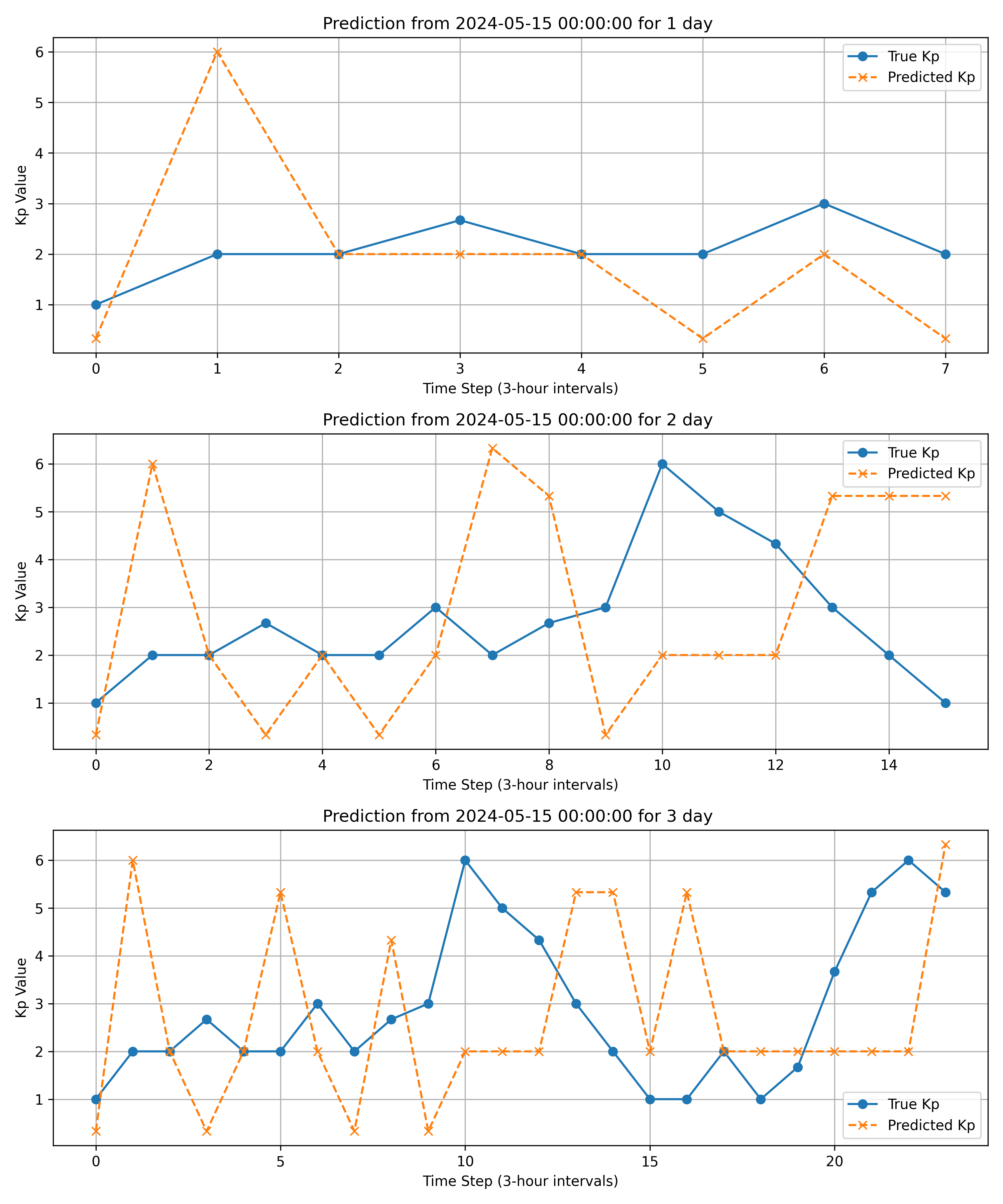}
        \caption{May 15}
        \label{fig:prediction_2024-05-15}
    \end{subfigure}
    
    \caption{\textbf{The multimodal model prediction results from May 12 to May 15}}
    \label{fig:predictions}
    
\end{figure}
\begin{figure}[htbp]   

    \vspace{0.5em}
    \begin{subfigure}[b]{0.475\textwidth}
        \centering
        \includegraphics[width=\textwidth]{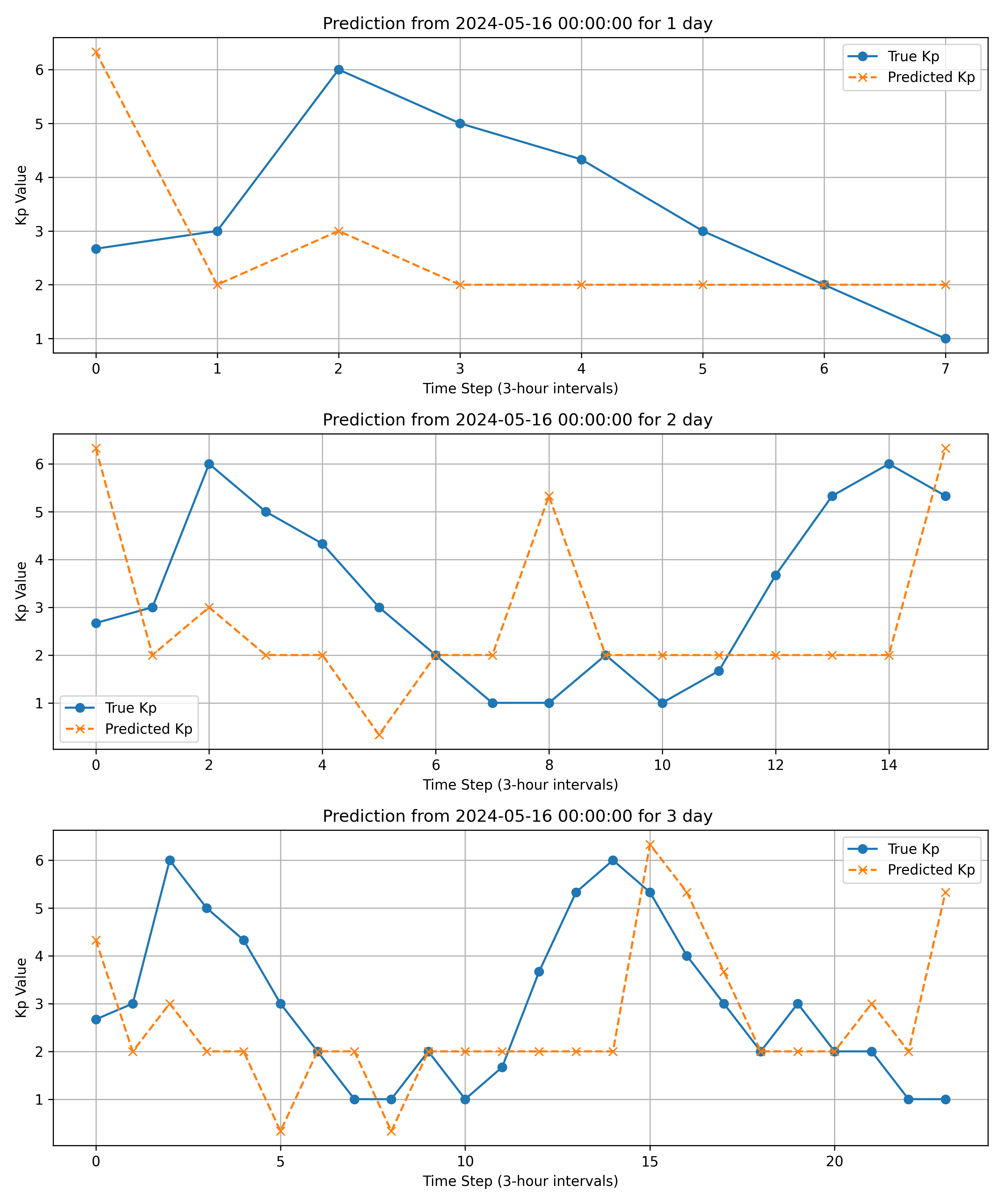}
        \caption{May 16}
        \label{fig:prediction_2024-05-16}
    \end{subfigure}

    \caption{\textbf{The multimodal model prediction results on May 16}}
    \label{fig:predictions}
    
\end{figure}
\begin{figure}[htbp]
    \centering
    \begin{subfigure}[b]{0.475\textwidth}
        \centering
          \includegraphics[width=\textwidth]{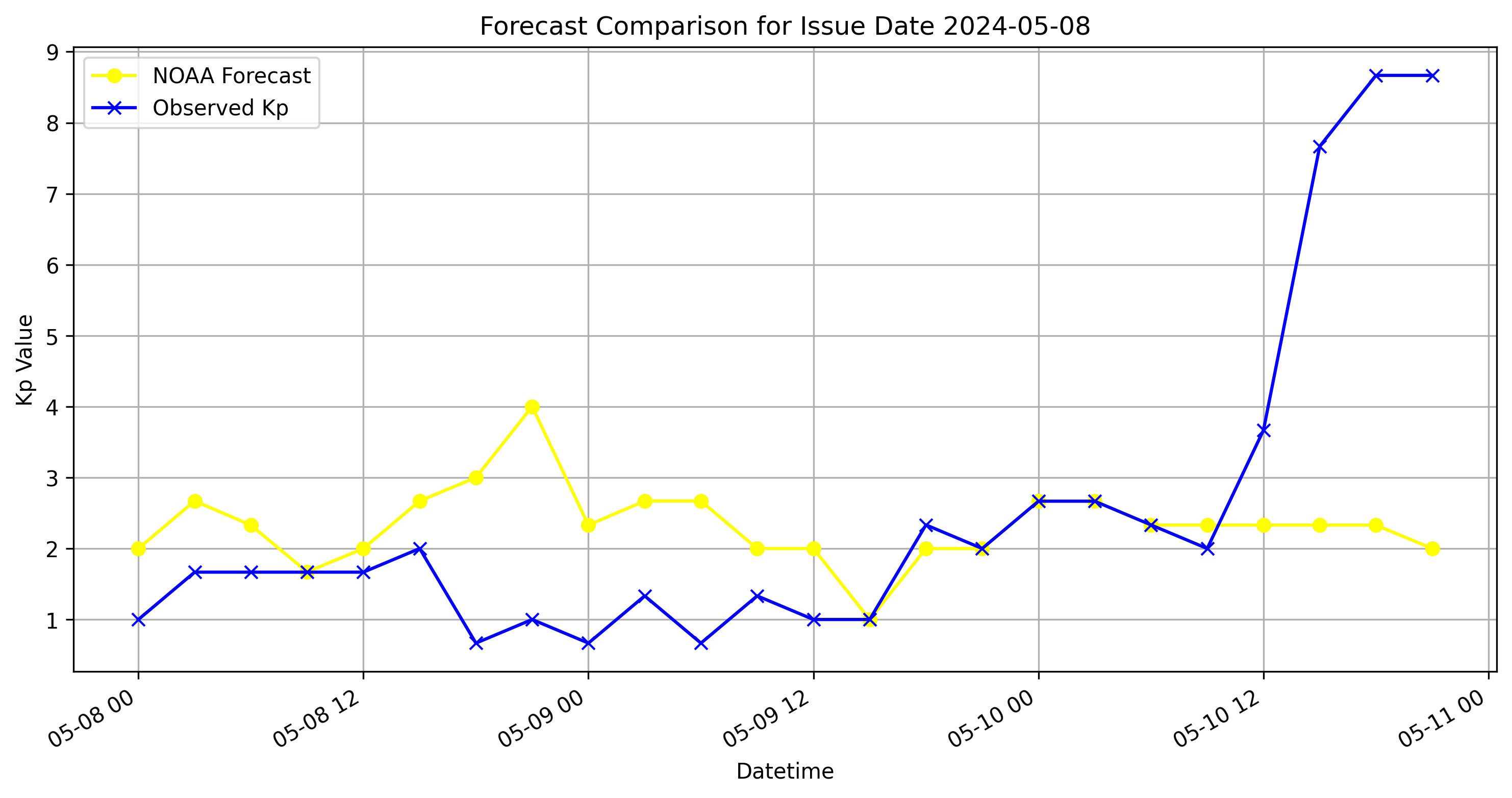}
        \caption{May 8}
         \label{fig:forecast_comparison_20240508}
    \end{subfigure}
    \begin{subfigure}[b]{0.475\textwidth}
        \centering
       \includegraphics[width=\textwidth]{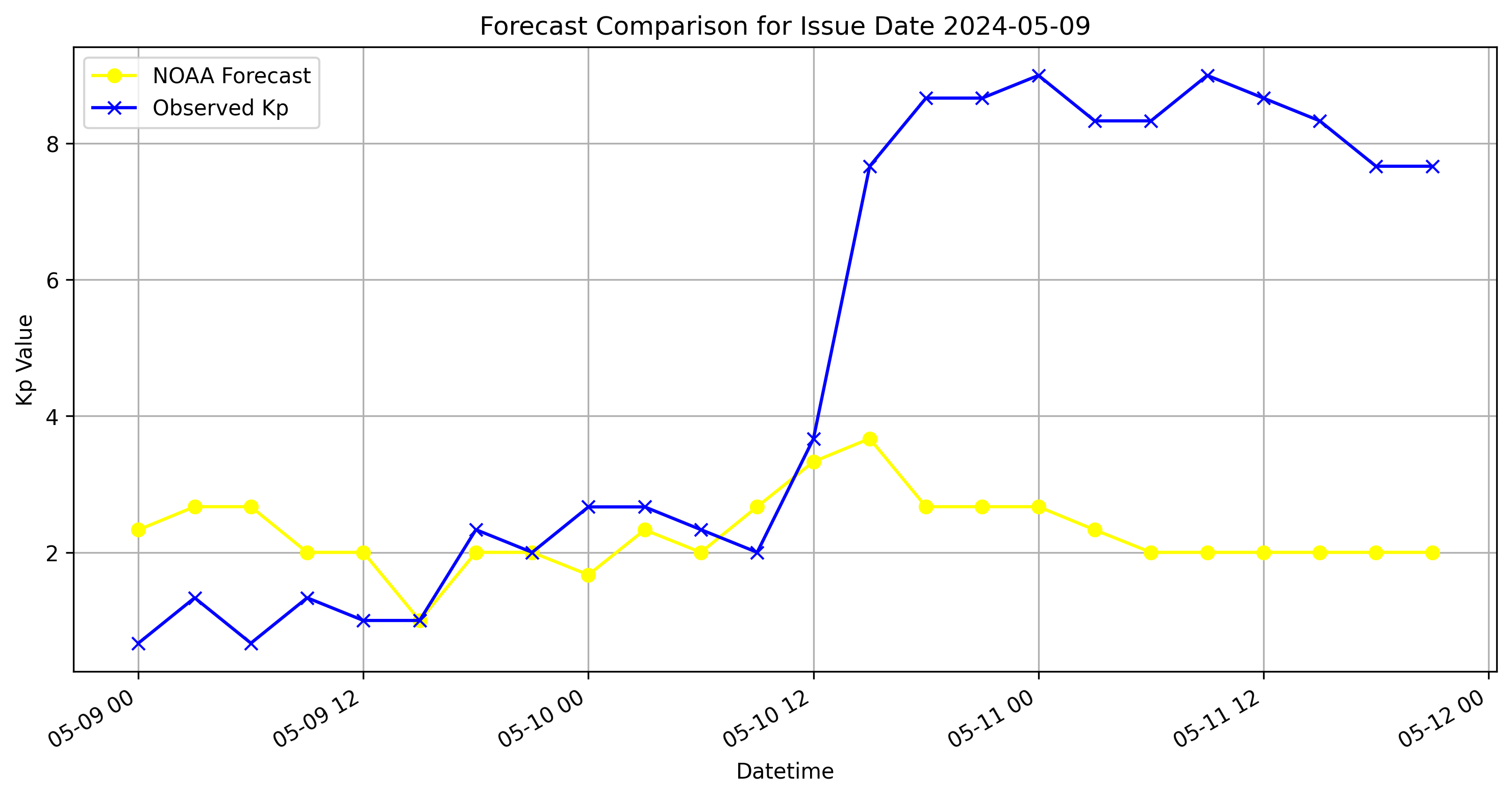}
        \caption{May 9}
        \label{fig:forecast_comparison_20240509}
    \end{subfigure}

    \caption{\textbf{The NOAA prediction results from May 8 to May 9}}
    \label{fig:predictions}
\end{figure}
\begin{figure}[htbp]
    \centering
    
    \vspace{0.5em}
    \begin{subfigure}[b]{0.475\textwidth}
        \centering
    \includegraphics[width=\textwidth]{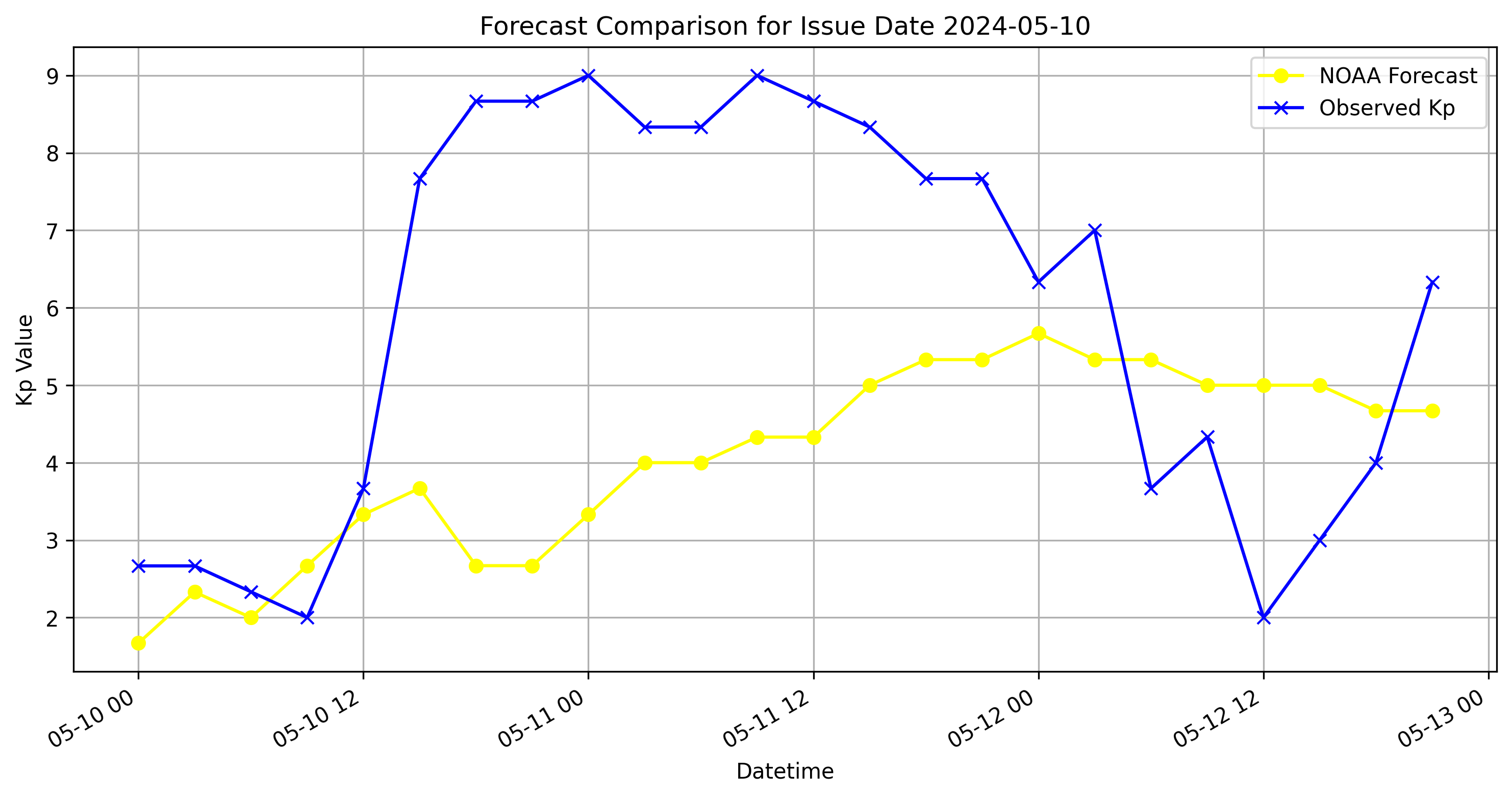}
        \caption{May 10}
        \label{fig:forecast_comparison_20240510}
    \end{subfigure}
    \begin{subfigure}[b]{0.475\textwidth}
        \centering
       \includegraphics[width=\textwidth]{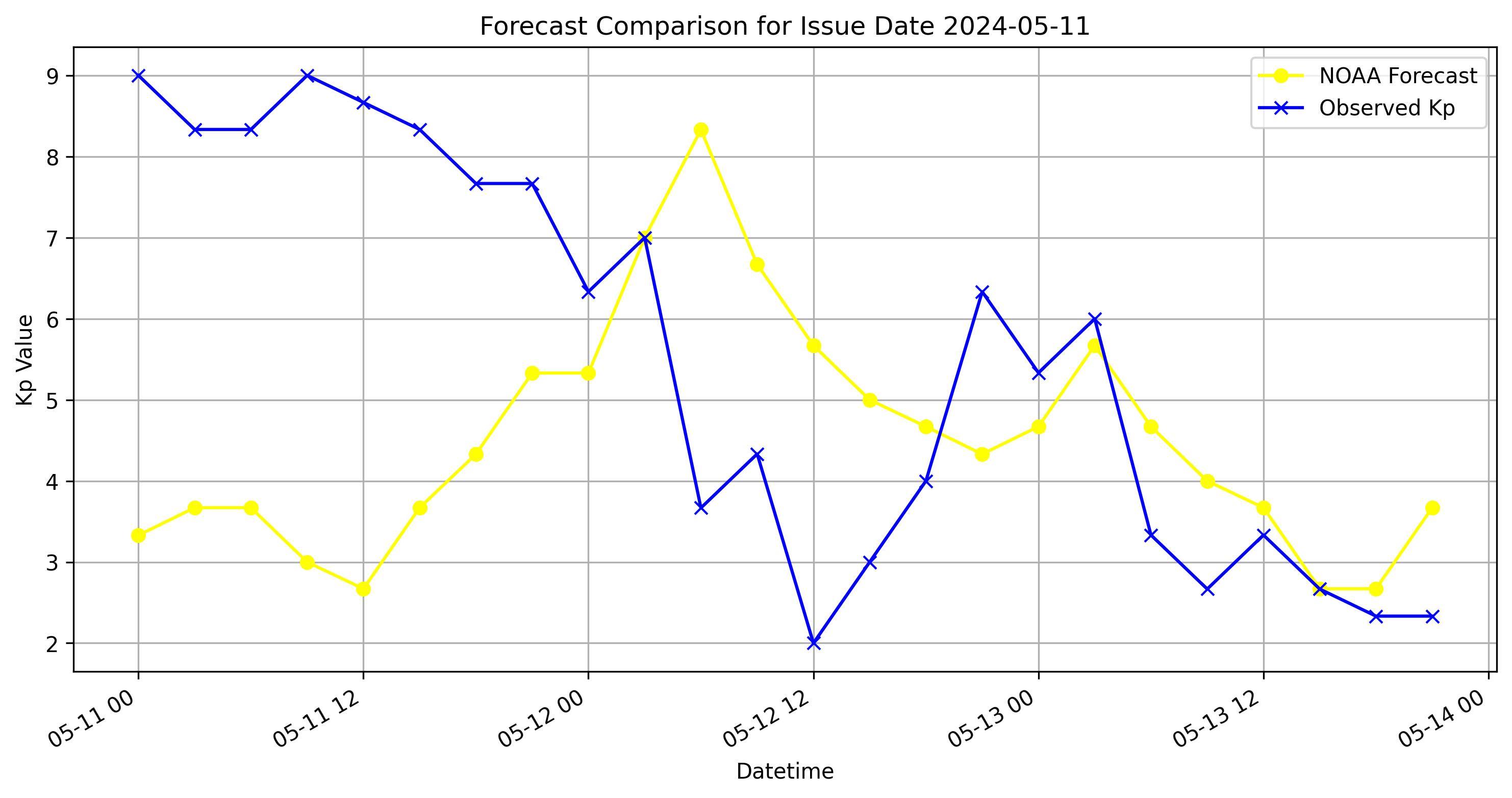}
        \caption{May 11}
        \label{fig:forecast_comparison_20240511}
    \end{subfigure}

   \vspace{0.5em}
    \begin{subfigure}[b]{0.475\textwidth}
        \centering
    \includegraphics[width=\textwidth]{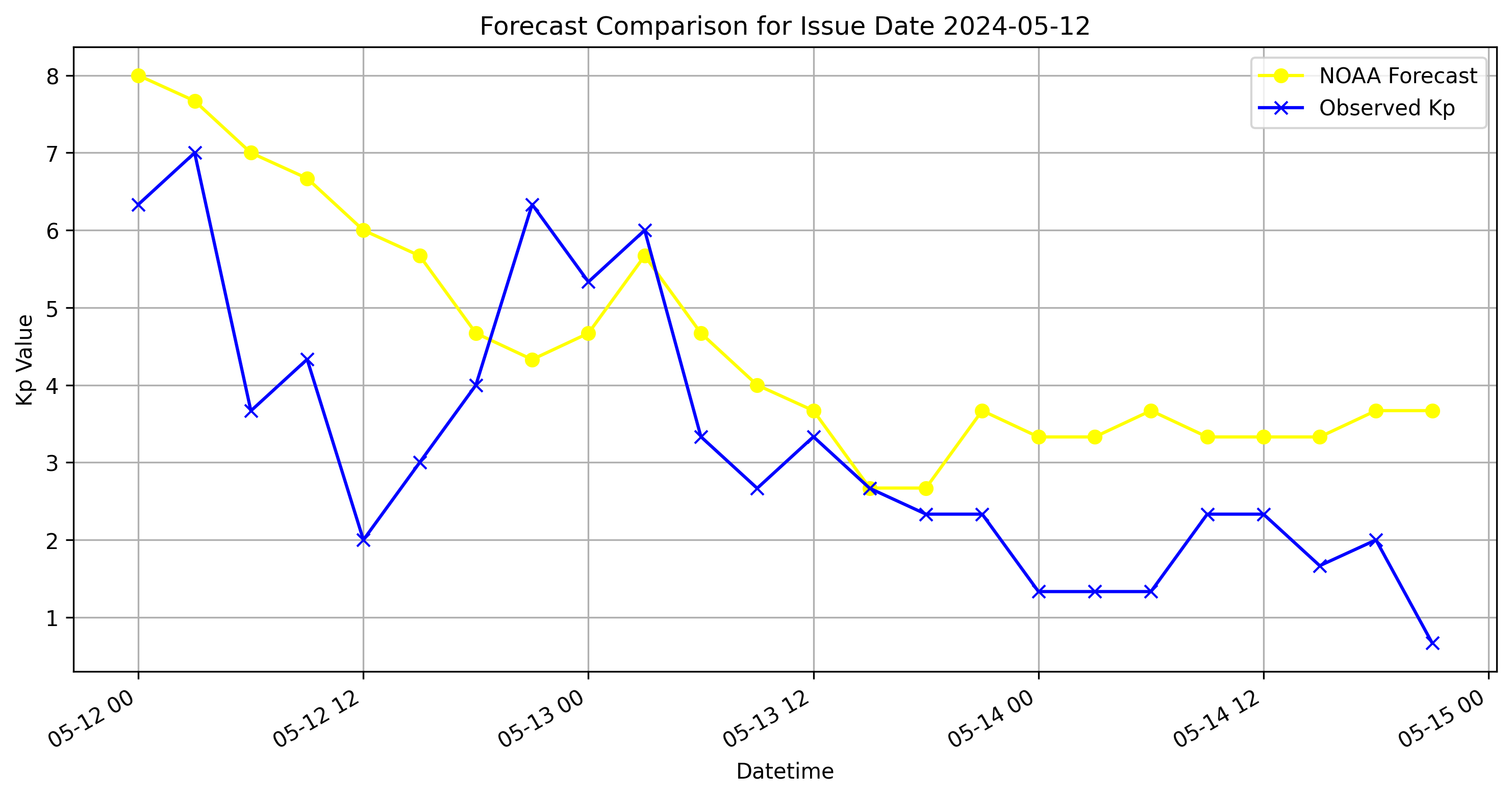}
        \caption{May 12}
        \label{fig:forecast_comparison_20240512}
    \end{subfigure}
    \begin{subfigure}[b]{0.475\textwidth}
        \centering
    \includegraphics[width=\textwidth]{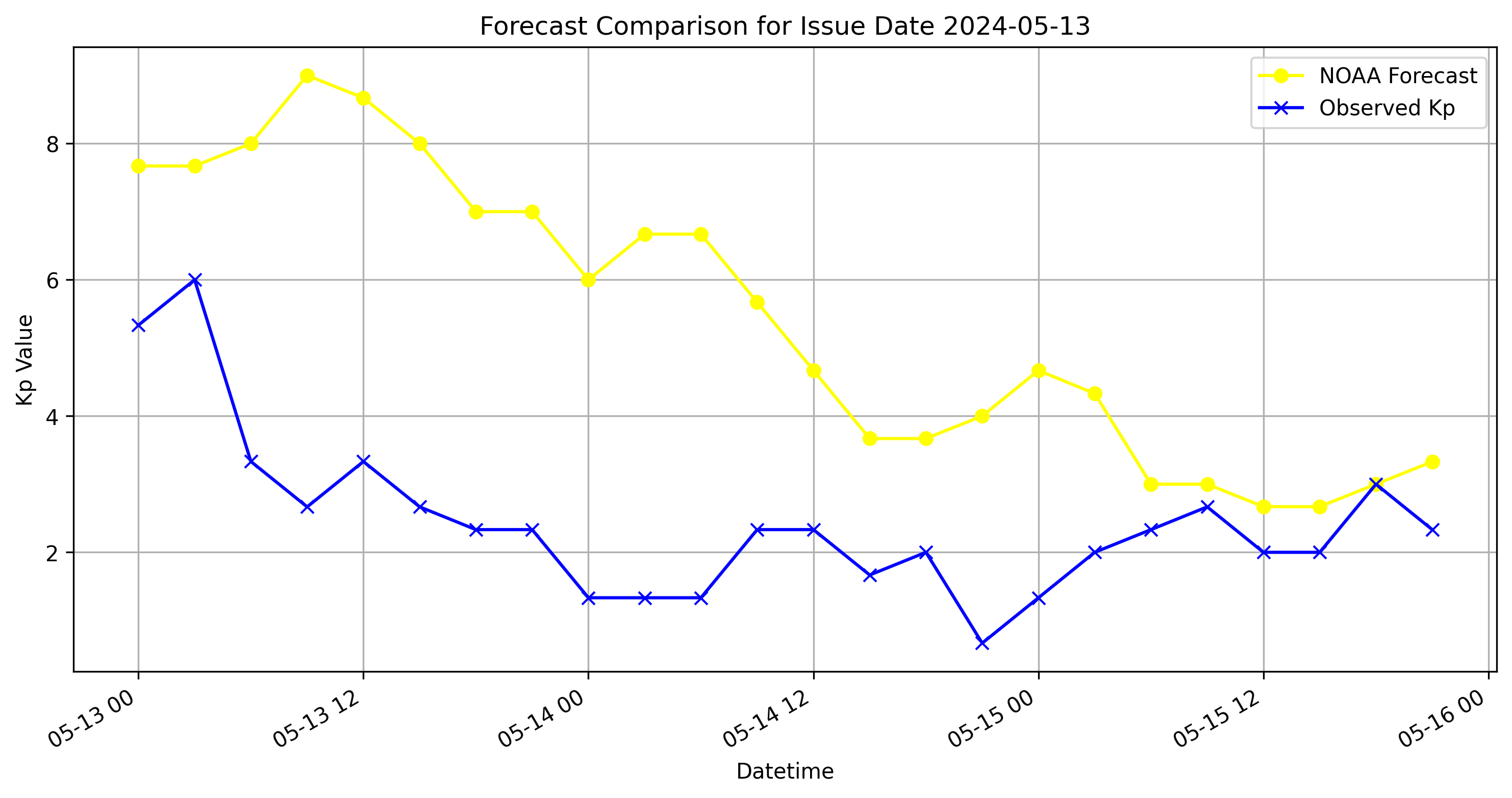}
        \caption{May 13}
        \label{fig:forecast_comparison_20240513}
    \end{subfigure}
    
    \vspace{0.5em}
    \begin{subfigure}[b]{0.475\textwidth}
        \centering
    \includegraphics[width=\textwidth]{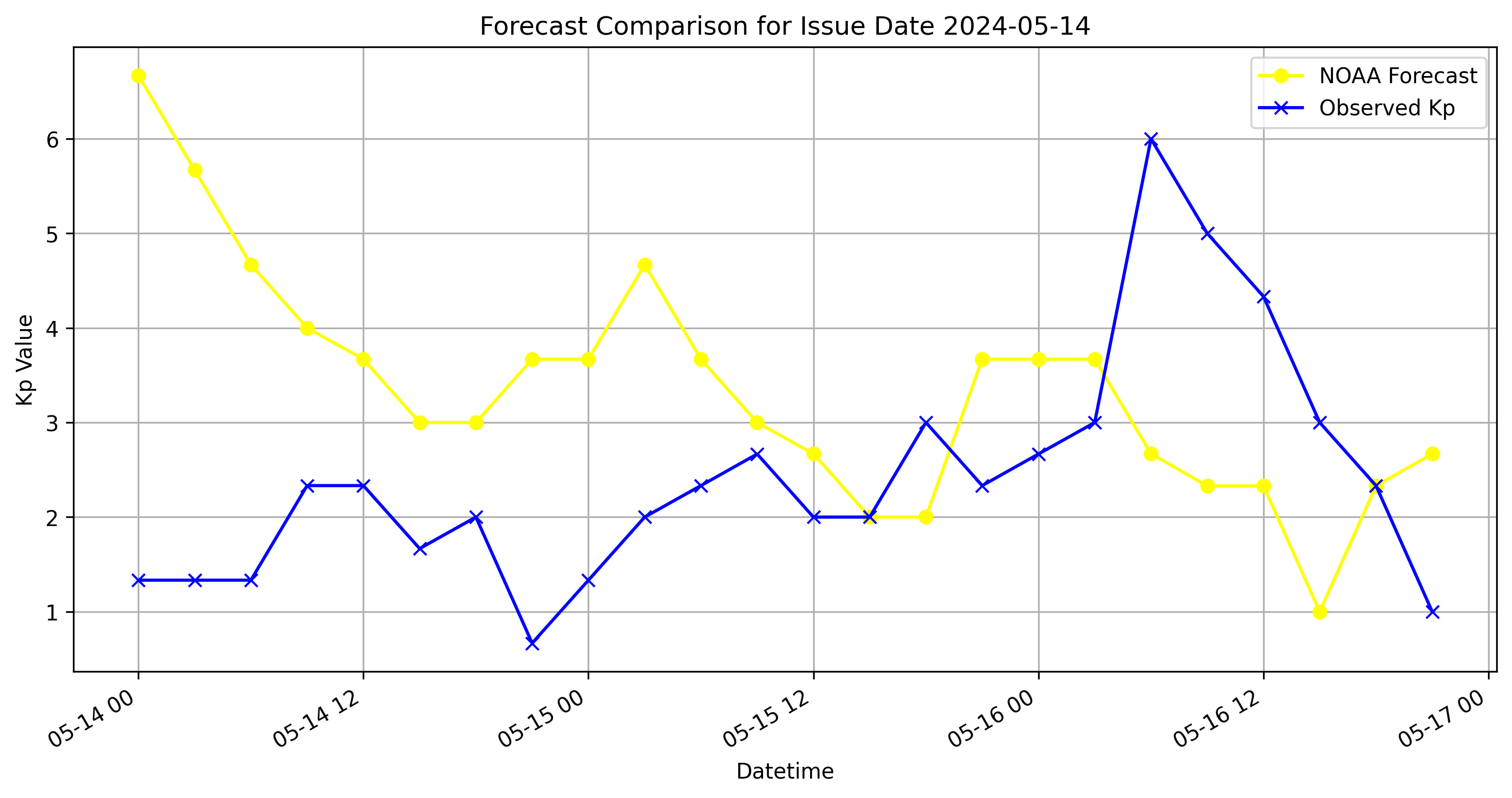}
        \caption{May 14}
        \label{fig:forecast_comparison_20240514}
    \end{subfigure}
    \begin{subfigure}[b]{0.475\textwidth}
        \centering
    \includegraphics[width=\textwidth]{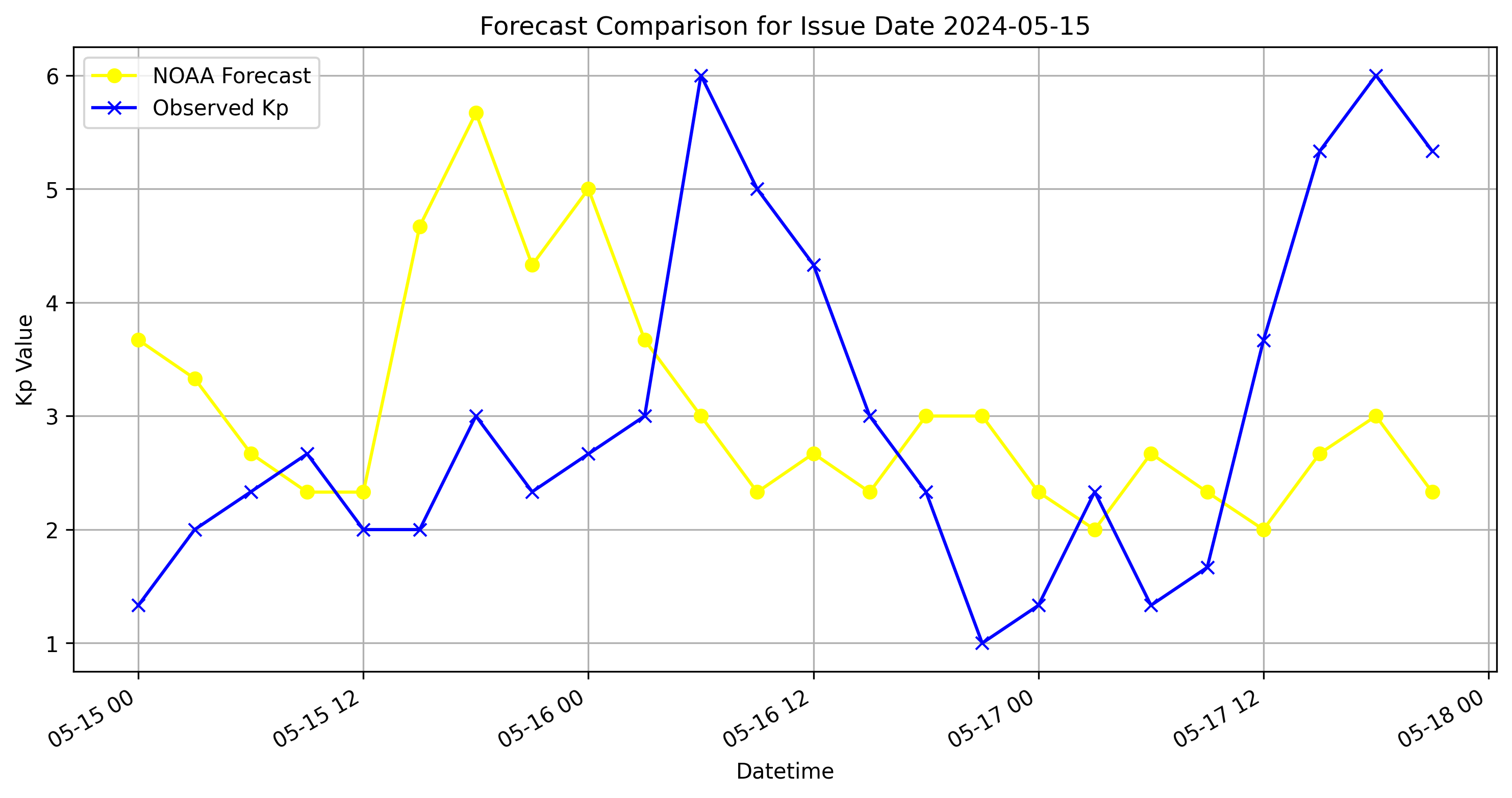}
        \caption{May 15}
        \label{fig:forecast_comparison_20240515}
    \end{subfigure}
    
    \vspace{0.5em}
    \begin{subfigure}[b]{0.475\textwidth}
        \centering
    \includegraphics[width=\textwidth]{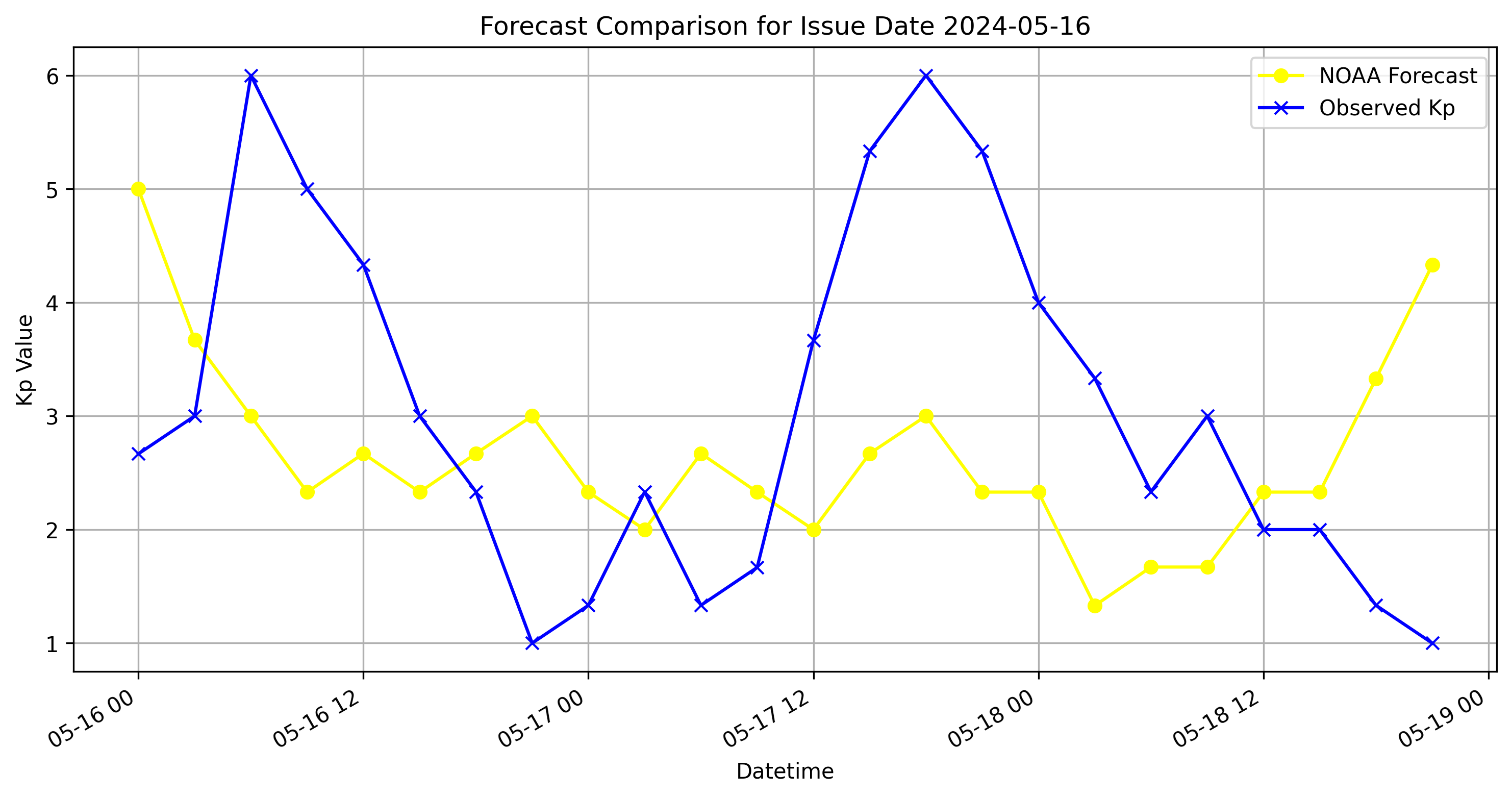}
        \caption{May 16}
        \label{fig:forecast_comparison_20240516}
    \end{subfigure}
      
    \caption{\textbf{The NOAA prediction results from May 10 to May 16}}
    \label{fig:predictions}
\end{figure}
\begin{figure}[htbp]
    \centering
    \begin{subfigure}[b]{0.475\textwidth}
        \centering
        \includegraphics[width=\textwidth]{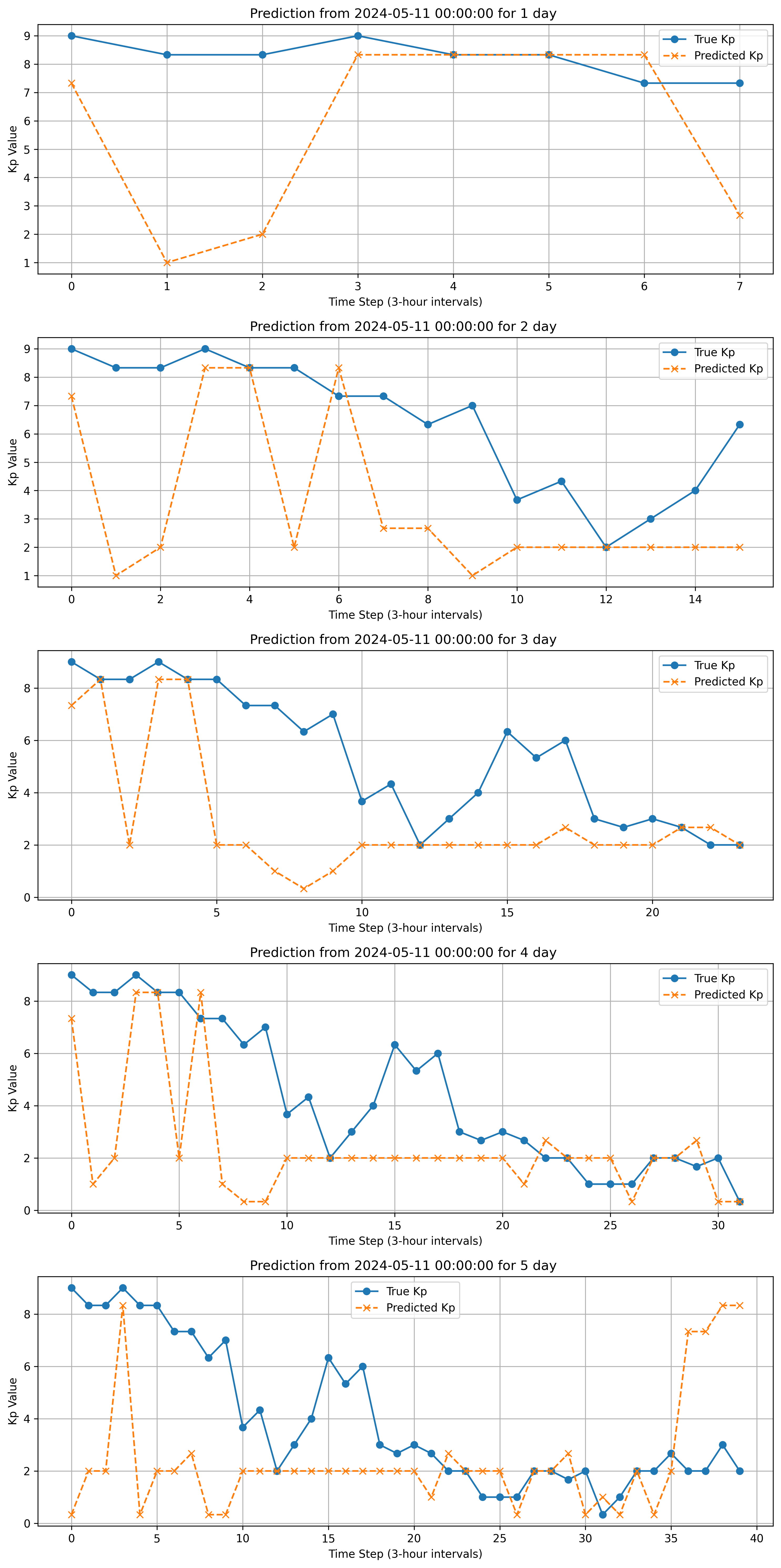}
        \caption{\textbf{The 5 days multimodal prediction on May 11}}
        \label{fig:prediction_20240511}
   \end{subfigure}
    \begin{subfigure}[b]{0.475\textwidth}
        \centering
        \includegraphics[width=\textwidth]{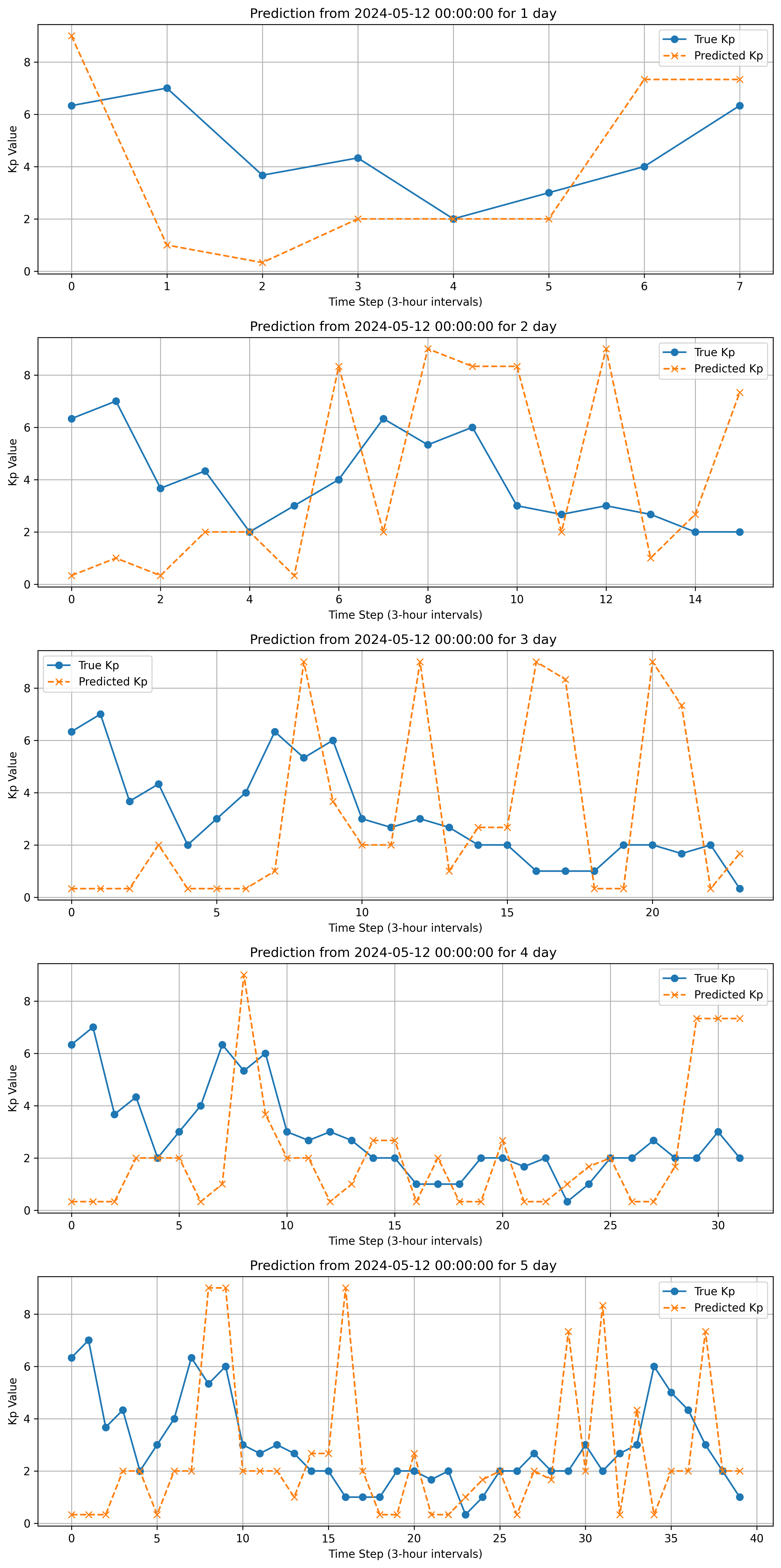}
        \caption{\textbf{The 5 days multimodal prediction on May 12}}
        \label{fig:prediction_20240512}
   \end{subfigure}
\end{figure}
\begin{figure}[htbp]
    \centering
    \includegraphics[width=\textwidth]{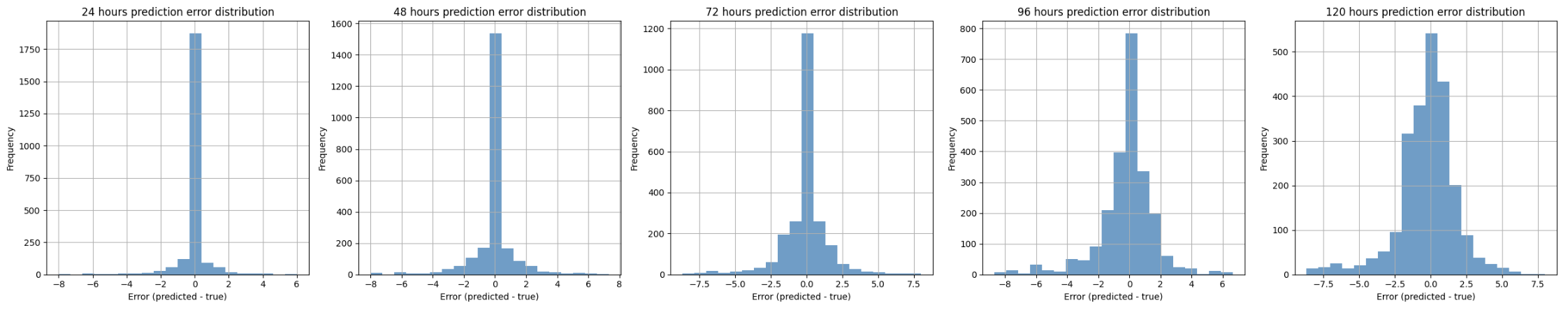}
    \caption{\textbf{The prediction error distribution}}
    \label{fig:distribution}

\end{figure}

Here are the results from both the multimodal model and NOAA. In which Figures 
\ref{fig:prediction_2024-05-08}, \ref{fig:prediction_2024-05-09}, \ref{fig:prediction_2024-05-10}, \ref{fig:prediction_2024-05-11}, \ref{fig:prediction_2024-05-12}, \ref{fig:prediction_2024-05-13}, \ref{fig:prediction_2024-05-14}, \ref{fig:prediction_2024-05-15}, and \ref{fig:prediction_2024-05-16} display the predictions from the multimodal model, while Figures 
\ref{fig:forecast_comparison_20240508}, \ref{fig:forecast_comparison_20240509}, \ref{fig:forecast_comparison_20240510}, \ref{fig:forecast_comparison_20240511}, \ref{fig:forecast_comparison_20240512}, \ref{fig:forecast_comparison_20240513}, \ref{fig:forecast_comparison_20240514}, \ref{fig:forecast_comparison_20240515}, and \ref{fig:forecast_comparison_20240516} show the corresponding forecasts from NOAA. The multimodal model provides one-day, two-day, and three-day predictions, which are crucial for identifying magnetic storms. A magnetic storm began on the afternoon of May 10, 2024 and persisted for several days. The multimodal model performed well on May 8, May 11, May 12, May 13, May 14, and May 15, but not on May 9 and May 10. Below, we compare the predictions from the multimodal model with those from the NOAA model.
The multimodal model is capable of making five-day predictions (see Figure \ref{fig:prediction_20240511}), and the error distribution is shown in Figure \ref{fig:distribution}.


\section{Conclusion}
This work presents a novel approach that integrates the Wasserstein distance into a Transformer based model for KP prediction, which integrating heterogeneous data sources including satellite measurements, solar images, and KP time series. A sliding window strategy is employed alongside advanced feature transformations to generate balanced training samples while preventing data leakage. By incorporating the Wasserstein distance into transformer and the loss function, our model effectively aligns the probability distributions across different modalities, leading to more consistent and robust representations. An additional binary branch for high KP detection further improves the model's ability to capture critical storm conditions. The model is continuously adapted using an online fine-tuning strategy on daily test data, which ensures that the forecasting remains responsive to local temporal dynamics and improves multi-step prediction accuracy.

Experimental comparisons with the NOAA model demonstrate our performance capturing both quiet periods and storm events. Future research will focus on exploring additional techniques and alternative formulations of the alignment loss to further improve model generalization and forecasting precision.

%
\bibliography{sc}
\bibliographystyle{sciencemag}

%
%
%
%
%
%


\section*{Acknowledgments}
We gratefully acknowledge the data support provided by several online resources. In particular, we thank the NASA Solar Dynamics Observatory for supplying the 193 Å solar images \cite{sdoData}, the Omniweb portal for integrating solar wind and near-Earth space environment data from multiple satellites \cite{omniweb}, and NOAA for providing the planetary Kp index data  \cite{noaaKp}. Without the generous data sharing by these organizations, this research would not have been possible.

\end{document}